\pdfoutput=1
\documentclass[10pt, logo, copyright, nonumbering]{nvidiatechreport}

\usepackage{booktabs}
\PassOptionsToPackage{numbers,sort&compress}{natbib}

\usepackage{url}
\usepackage{colortbl}
\usepackage{nicefrac}       %
\usepackage{microtype}      %
\usepackage[dvipsnames]{xcolor}         %
\usepackage{adjustbox}
\usepackage{xcolor}
\usepackage{subcaption}    %
\usepackage{float}
\usepackage{stfloats}
\usepackage{listings}
\usepackage{mdframed}
\usepackage{arydshln}
\usepackage{color,soul}
\usepackage{makecell}

\usepackage{amsmath,amssymb,stmaryrd,mathtools}
\SetSymbolFont{stmry}{bold}{U}{stmry}{m}{n}
\usepackage{xspace}
\usepackage{bbm}
\usepackage{mathtools}
\usepackage{bm}
\usepackage{graphicx}
\usepackage{lipsum}
\usepackage{wrapfig}
\usepackage{caption}
\usepackage{multicol, multirow}
\usepackage[table,dvipsnames]{xcolor}
\usepackage{siunitx}
\usepackage{enumitem}
\usepackage{authblk}
\usepackage{hyperref}
\usepackage{cleveref}
\usepackage{booktabs}
\usepackage[most]{tcolorbox}
\usepackage{tabularx}
\usepackage[table]{xcolor}
\usepackage{array}

\usepackage[numbers]{natbib}
\usepackage{chapterbib}

\tcbset{
  colback=gray!5,    
  colframe=black,    
  fonttitle=\bfseries,
  boxrule=0.5pt,
  arc=2mm,
  left=1mm,
  right=1mm,
  top=1mm,
  bottom=1mm,
}
\setlipsum{%
  par-before = \begingroup\color{gray},
  par-after = \endgroup
}

\definecolor{orange(sae/ece)}{rgb}{1.0, 0.49, 0.0}
\definecolor{teal(sae/ece)}{rgb}{0, 0.47, 0.52}
\definecolor{purple}{rgb}{0.74, 0.65, 1.0}
\definecolor{light_gray}{rgb}{0.9, 0.9, 0.9}
\definecolor{medium_gray}{rgb}{0.6, 0.6, 0.6} 
\definecolor{dark_gray}{rgb}{0.2, 0.2, 0.2} 
\definecolor{dark_blue}{rgb}{0.098, 0.239, 0.52}
\definecolor{dark_brown}{rgb}{0.3255, 0.004, 0.001}
\definecolor{dark_purple}{rgb}{0.478, 0.1569, 0.4863}
\definecolor{light_blue}{rgb}{0.33, 0.80, 1}
\definecolor{pastelblue}{RGB}{173,216,230}
\definecolor{pastelyellow}{RGB}{255,253,208}
\definecolor{pastelpink}{RGB}{255,209,220}
\definecolor{pastelgreen}{RGB}{176,226,172}
\definecolor{pastellavender}{RGB}{230,230,250}

\definecolor{NvidiaGreen}{RGB}{118, 185, 0}
\sethlcolor{red!15}

\usepackage{expl3}
\ExplSyntaxOn
\newcommand\latinabbrev[1]{
  \peek_meaning:NTF . {
    #1\@}%
  { \peek_catcode:NTF a {
      #1.\@ }%
    {#1.\@}}}
\ExplSyntaxOff

\usepackage{amsmath}

\usepackage{graphics} 
\usepackage{epsfig} 
\usepackage{mathptmx} 
\usepackage{times} 
\usepackage{amsmath} 
\usepackage{amssymb}  
\usepackage{multirow}
\usepackage{marvosym}
\usepackage{booktabs}
\usepackage{enumitem}
\usepackage{mathtools}
\usepackage{amsthm}
\usepackage{tabularx,float}
\usepackage{wasysym}
\usepackage[table]{xcolor}
\usepackage{hyperref}
\usepackage{url}
\usepackage[most]{tcolorbox}
\usepackage{graphicx}
\usepackage{capt-of}
\usepackage{caption}
\usepackage{anyfontsize}

\tcbuselibrary{listings,breakable}

\title{
\makebox[\textwidth][c]{
\parbox{0.95\textwidth}{
\centering
Physical Words: A Vocabulary for Structuring Latent Reasoning in Robot Manipulation
}
}
}

\author{
\parbox{\textwidth}{
\centering
\small
\mbox{Jiaming Liu$^{*,1,2}$},
\mbox{Xinghao Zhu$^{*,1}$},
\mbox{Yiming Zhang$^{*,2}$},
\mbox{Shanghang Zhang$^{*,2}$},
\mbox{Yinxi Wang$^{2}$},
\mbox{Yichu Yang$^{1}$},
\\
\small
\mbox{\textbf{Zhuoyang Liu$^{2}$}},
\mbox{\textbf{Dantong Niu$^{3}$}},
\mbox{\textbf{Soha Pouya$^{1}$}},
\mbox{\textbf{Juekun Li$^{1}$}},
\mbox{\textbf{Yan Chang$^{1}$}}

{\normalfont\small \textsuperscript{1} NVIDIA}
{\normalfont\small \textsuperscript{2} Peking University}
{\normalfont\small \textsuperscript{3} UC Berkeley}

{\normalfont\small
$^{*}$\textit{Co-First Authors}

}
}
}

\begin{document}

\begin{abstract}

{%
    \centering
     \vspace{-0.3cm}
    \includegraphics[width=\linewidth]{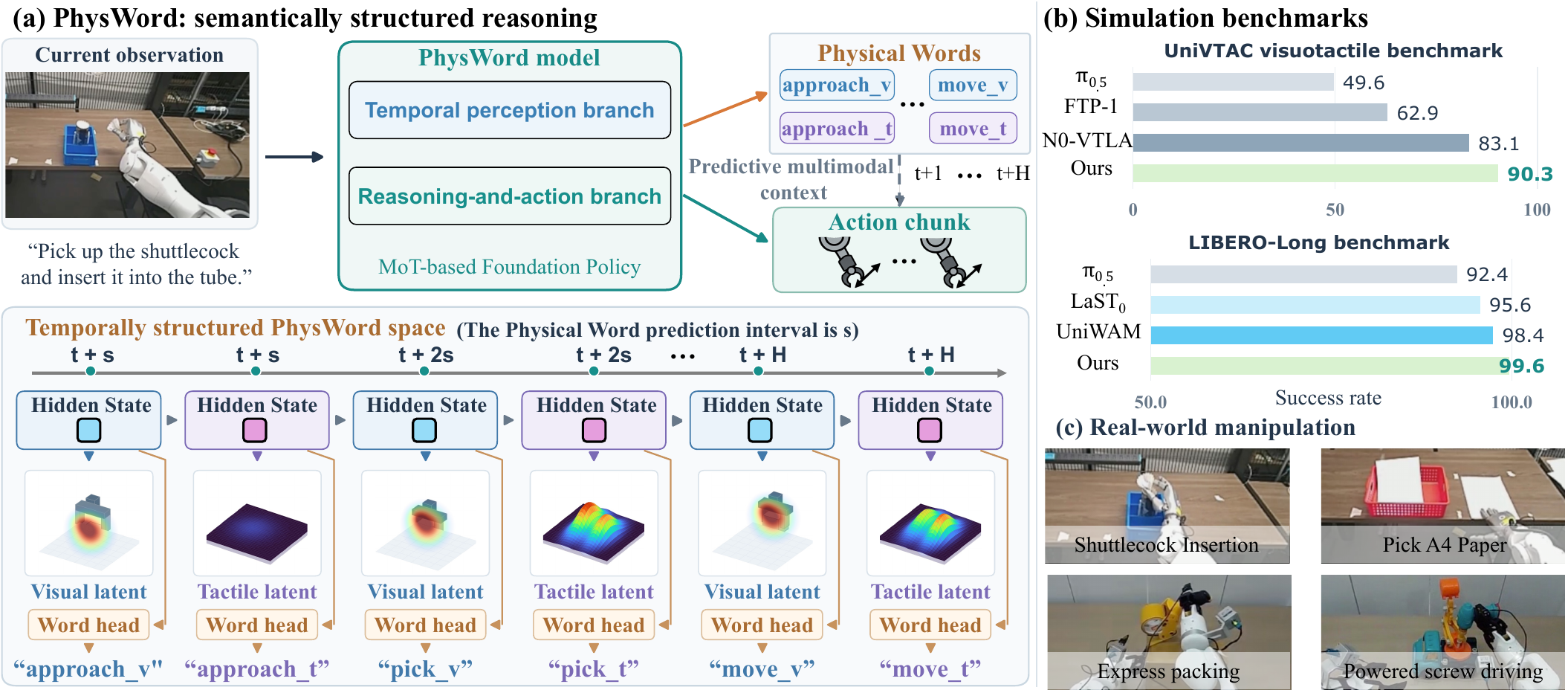}
    \par
    \captionsetup{
        font={normalfont,small},
        justification=justified,
        singlelinecheck=false
    }
    \vspace{-0.3cm}
    \captionof{figure}{%
        \textbf{Overview of PhysWord.}
        (a) PhysWord anchors continuous physical dynamics to atomic action words, with separate visual and tactile variants learned through direct latent supervision. Autoregressively generating Physical Words within a temporally structured space provides multimodal predictive context for action generation.
        (b) PhysWord achieves SOTA performance on both the visuotactile and vision-based simulation benchmarks.
        (c) Real-world experiments span bimanual gripper and dexterous-hand systems, both equipped with tactile sensing.
        Project page: \href{https://nvidia-isaac.github.io/video_to_data/physicalword/}{https://nvidia-isaac.github.io/video$\_$to$\_$data/physicalword/}.
    }
    \label{fig:teaser}
    \par
    \vspace{+0.3cm}
}

\medskip

\noindent\textbf{Abstract:}
Multimodal reasoning provides robotic foundation models with predictive context needed for reliable action generation. However, language-based reasoning captures task semantics but struggles to express fine-grained physical dynamics, while latent reasoning models future state evolution but offers limited structure for organizing these dynamics into reusable reasoning primitives. To bridge this gap, we introduce \textbf{PhysWord}, a reasoning-before-acting foundation model built on \textbf{Physical Words}, which use atomic action verbs as shared semantic anchors for continuous representations of physical dynamics. Specifically, PhysWord predicts upcoming atomic action words and directly aligns their contextual hidden states with latent targets derived from the corresponding future observations. Building on these primitives, we construct a temporally structured PhysWord space with visual and optional tactile variants of each action word. Autoregressively predicting Physical Words across future timesteps captures transitions between execution stages, providing structured multimodal context for action generation. PhysWord achieves state-of-the-art success rates of 90.3\% on visuotactile UniVTAC and 99.6\% on vision-based LIBERO-Long, and outperforms competing methods on real-world tasks using bimanual grippers and dexterous hands. Further evaluations demonstrate that our structured latent reasoning framework generalizes to unseen scenarios and enables long-horizon manipulation.

\end{abstract}

\maketitle

\section{Introduction} 
\label{sec:introduction}

Vision-Language-Action (VLA) models have emerged as a promising foundation for general-purpose robotic manipulation~\cite{kim2024openvla,liu2025hybridvla}, integrating the semantic knowledge of pretrained vision-language models (VLMs)~\cite{karamcheti2024prismatic, bai2025qwen3, beyer2024paligemma} with robot action learning. Reliable manipulation, however, requires anticipating how the physical environment evolves as the robot acts. 
This need has motivated reasoning-before-acting VLA approaches~\cite{huang2025thinkact,liu2026last}, which generate intermediate task plans or predictive context to guide action generation.
World Action Models (WAMs) further integrate future world prediction with action generation, enabling policies to leverage dynamics knowledge learned from video and manipulation data. These advances highlight the value of multimodal reasoning in providing predictive context that aligns robot actions with evolving environments.

Existing reasoning approaches differ in how they represent predictive context.
On the one hand, some VLA models adopt the reasoning paradigm of pretrained VLMs, using language to decompose tasks and express task-level action semantics~\cite{intelligence2025pi05visionlanguageactionmodelopenworld, lin2025onetwovla, zawalski2025roboticcontrolembodiedchainofthought}. However, linguistic descriptions remain physically coarse, as a verb such as ``rotate'' identifies an operation without fully conveying its fine-grained dynamics.
On the other hand, WAMs~\cite{ye2026world, wang2026openwam} and latent reasoning methods~\cite{liu2026last, bi2025motus, tian2025interndata} model future state evolution, providing rich predictive representations of interaction dynamics. Yet these approaches do not necessarily organize their representations around shared action semantics. Since the same manipulation skill can be applied across different objects and scenes, a useful reasoning representation should preserve shared semantics while capturing context-specific physical dynamics. This raises a central question: Can atomic action semantics organize physical latent representations into reusable reasoning primitives?

To this end, we introduce PhysWord, a reasoning-before-acting foundation model built on Physical Words, which use atomic action verbs as semantic anchors for continuous latent dynamics. We assemble a vocabulary of atomic action verbs, such as “pick,” “rotate,” and “push,” based on the action taxonomy of prior work~\cite{yuan2026qwen, chen2025rh20t}.
As shown in Figure~\ref{fig:teaser}(a), PhysWord predicts upcoming atomic action words from the current observation context and directly aligns their contextual hidden states with latent targets derived from the corresponding future observations.
Joint semantic and latent supervision encourages these representations to capture shared action semantics while retaining action-related physical dynamics.
To capture how these dynamics evolve during execution, we organize Physical Words at multiple future timesteps within the action-chunk horizon, forming a temporally structured PhysWord space.
Within this space, we use separate visual and tactile variants of each action word to capture complementary motion and contact dynamics, including tactile variants only for tactile-enabled tasks.
By autoregressively predicting these Physical Words, PhysWord captures changes in action semantics and physical dynamics across execution stages, providing structured multimodal predictive context for long-horizon manipulation.
This token interface adapts to the available sensory modalities, bringing demonstrations with and without tactile observations into a unified training formulation. This flexibility enables us to jointly pretrain PhysWord on approximately 7,000 hours of real-robot and human-hand demonstrations.
Across this corpus, we use a VLM~\cite{openai2026gpt56} to automatically segment trajectories into atomic action stages and align the resulting atomic verbs with the corresponding observations.

As shown in Figure~\ref{fig:teaser}(b), PhysWord achieves success rates of 90.3\% on visuotactile UniVTAC and 99.6\% on vision-only LIBERO-Long, outperforming prior SOTA methods.
Real-world experiments on bimanual gripper and dexterous-hand platforms with visual and tactile sensing further demonstrate its effectiveness in contact-rich, precise, and long-horizon manipulation.
To assess reusability, we evaluate both unseen scene configurations and extended task sequences absent from training.
PhysWord outperforms competing methods under unseen backgrounds, object instances, and object positions on \textit{Powered Screw Driving} and \textit{Pick A4 Paper}, and successfully executes extended manipulation sequences, supporting reuse across both scenarios and task stages.
To assess the predictive value of its tactile representations, we replace measured tactile inputs with predicted latents at inference.
Remarkably, the resulting policies retain strong performance with only modest degradation, suggesting that semantic structuring preserves contact information needed for control and enables deployment without online tactile sensing.
\textbf{Our contributions:}

\begin{itemize}[nosep, leftmargin=*]

\item We introduce PhysWord, a reasoning-before-acting foundation model built on Physical Words, which use atomic action verbs as semantic anchors for continuous physical dynamics by directly supervising their contextual hidden states with future latent targets.

\item We construct a temporally structured PhysWord space with visual and tactile variants of atomic action words, capturing evolving motion and contact dynamics for action generation.

\item PhysWord achieves 90.3\% on UniVTAC and 99.6\% on LIBERO-Long, outperforming prior SOTA methods. Experiments on real-world tasks demonstrate superior performance, stronger generalization, and long-horizon manipulation through sequential reuse of Physical Words.

\end{itemize}

\section{Related Work}
\label{sec:related}


\textbf{Vision-Language-Action (VLA) Models.}
VLA models~\cite{kim2024openvla, black2024pi_0, li2023manipllmembodiedmultimodallarge, bjorck2025gr00t, liu2025hybridvla, wen2025dexvla} adapt pretrained VLMs to generate robot actions from visual observations and language instructions.
Reasoning-before-acting VLAs introduce intermediate reasoning to anticipate task progression.
Language-based approaches, including subtask prediction, embodied chain-of-thought (CoT), and motion descriptions~\cite{intelligence2025pi05visionlanguageactionmodelopenworld, lin2025onetwovla, zawalski2025roboticcontrolembodiedchainofthought, belkhale2024rthactionhierarchiesusing}, convey task semantics and subtask plans but struggle to express fine-grained dynamics.
Beyond language, recent methods predict future visual observations~\cite{wang2025unifiedvisionlanguageactionmodel, zhao2025cot, chen2025unifieddiffusionvlavisionlanguageaction, gao2025adaworld, bu2025agibot, zhang2025dreamvla} or incorporate anticipated signals from multiple modalities~\cite{liu2025mla, cen2025worldvla, gu2025manualvla}.

\textbf{Latent Reasoning and World Action Models (WAMs).}
Latent CoT represents intermediate reasoning through continuous latent states in LLMs and VLMs~\cite{hao2024training, deng2024explicit, yang2025machine, wang2025monet}.
ATLAS~\cite{guo2026atlas} further highlights the importance of LLMs' native language-based CoT for general reasoning tasks.
In robotics, recent VLA models adopt latent reasoning to guide control using compact representations of future physical dynamics~\cite{zhang2026atomicvla, huang2025thinkact, tan2025latent}.
Subsequent work incorporates reinforcement learning~\cite{chen2026lastr1} and latent alignment between human and robot demonstrations~\cite{liu2026lasthd}.
Related WAMs jointly model video and action sequences~\cite{kim2026cosmos, agarwal2026cosmos, ye2026world, wang2026openwam}.
Several approaches condition action generation on latent representations of predicted future states~\cite{hu2024video, yuan2026fast, luo2026being, wang2026tacdyn, chen2026uniwam}.
Although these approaches capture rich future dynamics, their latent representations are not structured around shared action semantics.
PhysWord introduces this structure by anchoring future dynamics to atomic action verbs, forming reusable reasoning primitives that provide predictive context for action generation.

\section{Methodology}
\label{sec:method}

In this section, we present PhysWord for structured latent reasoning in robotic manipulation, including its architecture (Section~\ref{sec:architecture}), Physical Word formulation (Section~\ref{sec:physical_words}), temporally structured PhysWord space (Section~\ref{sec:physword_space}), and training objectives (Section~\ref{sec:training_objectives}).

\subsection{Preliminaries}
\label{sec:preliminaries}

\textbf{Problem Formulation.} We consider language-conditioned robotic manipulation with visual observations and optional tactile sensing. At time $t$, the robot receives a language instruction $\ell$, multi-view visual observations $\mathcal{V}_t$, and, when available, tactile observations $\mathcal{T}_t$. We denote the current observations and their temporal history by $\mathcal{O}_t$. Conditioned on this context, the policy predicts an action chunk:
$
    \mathbf{A}_t
    = (\mathbf{a}_{t+1},\ldots,\mathbf{a}_{t+H})
    \sim \pi_{\theta}(\cdot \mid \mathcal{O}_t,\ell),
$
where $H$ denotes the action prediction horizon and $\mathbf{a}_t$ is an embodiment-specific control vector.
For dual-arm gripper platforms, each action is $14$-DOF, comprising a 3-DOF relative translation, a 3-DOF Euler-angle rotation, and a binary gripper command for each arm. For dexterous-hand platforms, each action is $56$-DOF, comprising $12$ end-effector pose components across the two arms and $44$ joint commands across the two Sharpa hands.

\subsection{PhysWord Model Architecture}
\label{sec:architecture}
\begin{figure}[t]
    \centering
    \includegraphics[width=\textwidth]{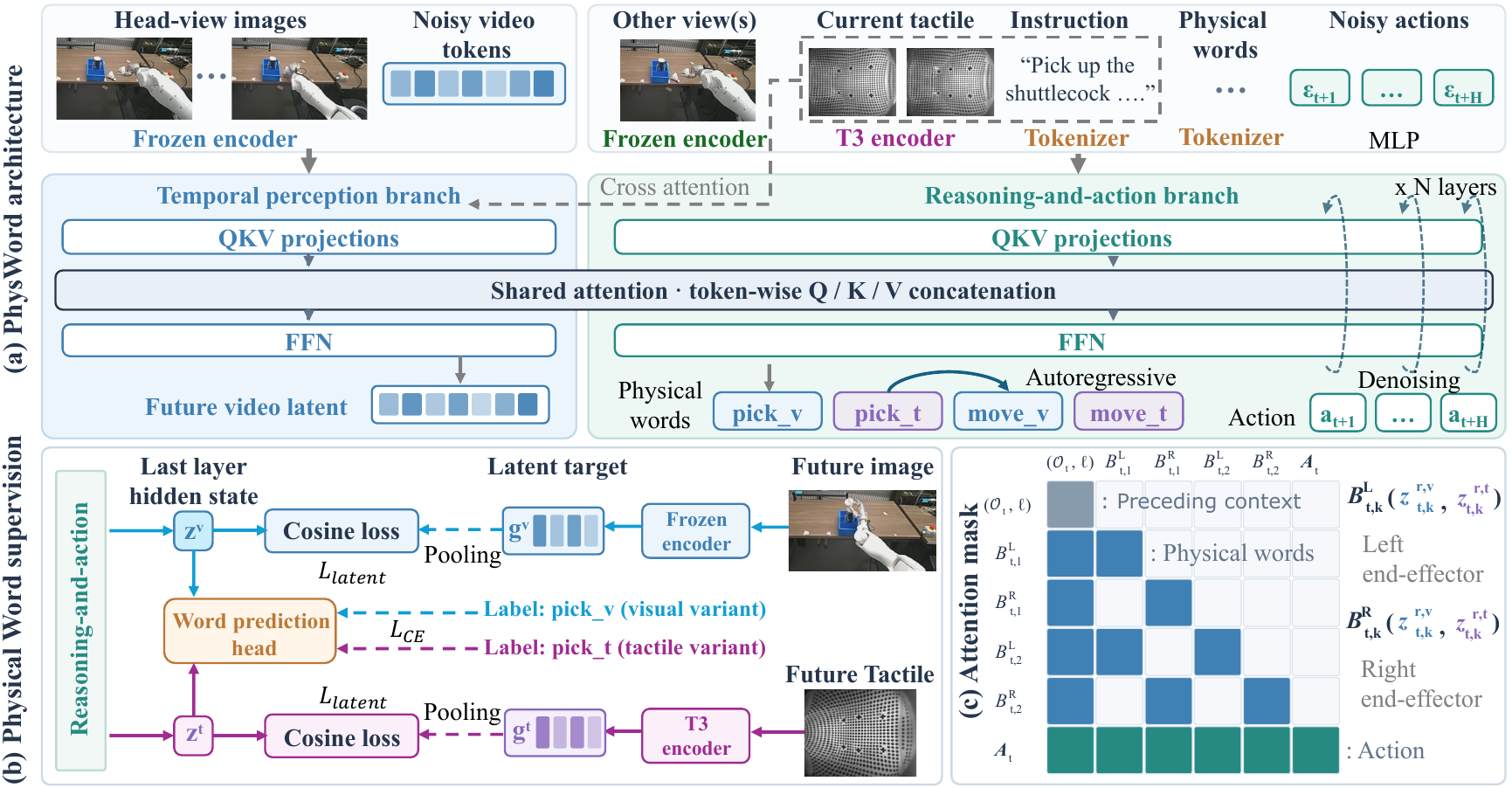} 
    \vspace{-0.5cm}
    \caption{\textbf{PhysWord framework.}
(a) PhysWord connects temporal perception and reasoning-and-action branches through shared attention in a MoT architecture, generating Physical Words autoregressively before denoising actions. (b) The hidden states of visual and tactile Physical Word variants receive joint action-label cross-entropy supervision and cosine alignment with future sensory latent targets. (c) Physical Words use a blockwise-causal attention mask, with full attention within each end-effector’s timestep block, causal attention across its blocks, and no cross-effector attention.
    }
    \label{fig:method}
\end{figure}

PhysWord adopts a mixture-of-transformers (MoT) architecture with a temporal perception branch initialized from Cosmos-Predict2.5-2B~\cite{ali2025world} and a reasoning-and-action branch initialized from Qwen3-VL-2B~\cite{bai2025qwen3}. As shown in Figure~\ref{fig:method} (a), the temporal branch encodes visual history and anticipated scene evolution, while the reasoning-and-action branch generates Physical Words followed by robot actions. 

\textbf{Multimodal inputs.}
The temporal perception branch receives five head-view observations from $t-4$ to $t$, each at a resolution of $\texttt{256}\times\texttt{256}$, and predicts the next $H$ frames.
The reasoning-and-action branch processes the remaining camera views and the language instruction $\ell$, retaining adaptive-resolution image processing. Both visual encoders are frozen, while all other components remain trainable.
For tactile-enabled tasks, we use the T3 tactile encoder~\cite{zhao2024transferable}, initialized from FTP-1 pretraining, to extract a $512$-dimensional feature from each 3D visuotactile observation. We aggregate these features according to the end-effector configuration. 
For each two-finger gripper, we concatenate its two sensor features into a $1024$-dimensional vector per timestep, then concatenate the vectors at $t$ and $t-1$ to form a $2048$-dimensional tactile token. For dexterous hands, we concatenate five fingertip features into a $2560$-dimensional vector and apply channel-wise pooling to obtain the $2048$-dimensional token. Both configurations yield one tactile token per end-effector, matching the LLM's word embedding dimension and facilitating joint pretraining across grippers and dexterous hands. 
Bimanual tasks use two tactile input tokens.
Tactile features and language instructions condition the temporal perception branch through Cosmos's cross-attention and the reasoning-and-action branch through input tokens in the LLM's embedding space.

\textbf{Shared attention and action generation.}
The two branches retain independent QKV projections and FFNs, along with their original pretrained positional encodings. After projection, their query, key, and value features are concatenated along the token dimension for shared attention.
At inference, Cosmos performs a single denoising step with full attention to obtain predictive features, whose keys and values are cached for subsequent token generation.
As shown in Figure~\ref{fig:method}, PhysWord then autoregressively generates Physical Words under a blockwise-causal mask.
Noisy action tokens follow the Physical Word tokens within the reasoning-and-action branch. They can attend to all preceding multimodal context and are trained using the flow-matching objective $\mathcal{L}_{\mathrm{act}}$. Following~\cite{black2024pi_0}, a shared MLP maps action tokens to a unified action space with sufficient dimensions to accommodate the supported embodiments.

\subsection{Physical Words}
\label{sec:physical_words}

A Physical Word is a reasoning primitive that couples an atomic action word label with a contextual hidden state representing future visual or tactile dynamics.

\textbf{Atomic action vocabulary.}
Large-scale language pretraining provides LLMs with rich semantic knowledge, while native next-token prediction offers a natural interface for expressing intermediate reasoning~\cite{guo2026atlas}. In robotic manipulation, continuous latents compactly encode multimodal context, but representations of the same action across scenarios may also capture incidental visual variations. Therefore, we introduce Physical Words to organize latent reasoning around atomic action semantics, preserving shared skill semantics while capturing context-dependent details. 
Based on prior action taxonomies~\cite{yuan2026qwen, chen2025rh20t, nasiriany2022augmenting}, we define $16$ action labels as semantic anchors for continuous latent representations. Additional action words can be incorporated by initializing their Physical Word variants from the corresponding pretrained LLM embeddings.
\begin{tcolorbox}[
    colback=gray!8,
    colframe=gray!25,
    boxrule=0.4pt,
    arc=2pt,
    left=6pt,
    right=6pt,
    top=5pt,
    bottom=5pt,
    fontupper=\small
]
\textbf{Atomic Action Vocabulary}\par\smallskip
Each label has a visual variant (\texttt{\_v}) and a tactile variant
(\texttt{\_t}), e.g., \texttt{pick\_v} and \texttt{pick\_t}.\par\smallskip
\textbf{Movement:} Approach, Move.
\textbf{Manipulation:} Pick, Place, Flip, Rotate, Toggle, Open,
Close, Push, Pull, Insert, Press, Click, Strike, Handover.
\end{tcolorbox}

\textbf{From action words to latent representations.}
Each Physical Word associates an action-word label $w$ with a contextual hidden state $\mathbf z$ from which the label is predicted. Let $e(w)$ denote the trainable vocabulary embedding of $w$ and $\mathbf g$ the corresponding future sensory latent target used to supervise $\mathbf z$.
To capture future physical dynamics, we predict both visual dynamics and tactile contact states. Since visual and tactile targets occupy distinct latent spaces and may impose conflicting supervision on a single hidden state, we instantiate separate visual and tactile variants of each action word. Both are initialized from the same pretrained embedding~\cite{bai2025qwen3} while maintaining independent trainable vocabulary entries. This preserves a common semantic initialization while allowing modality-specific hidden states to represent future dynamics.
For vision-only tasks, we omit tactile inputs and generate only visual Physical Words, such as \texttt{pick\_v}.
As shown in Figure~\ref{fig:method} (b), the word prediction head maps $\mathbf z$ to vocabulary logits, supervised by cross-entropy ($L_{\mathrm{CE}}$) against $w$. The same $\mathbf z$ is directly aligned with $\mathbf g$ using cosine distance ($L_{\mathrm{latent}}$). This joint supervision couples action semantics with future physical dynamics within a shared contextual representation.
PhysWord is trained with blockwise teacher forcing and generates Physical Words autoregressively over future timestep blocks at inference.
Compared with latent supervision alone, Physical Words yield lower intra-class distances among latent representations of the same action, and more focused attention on action-relevant regions, as shown in Section~\ref{sec:ablations} and Appendix~\ref{app:representation_analysis}.
Appendix~\ref{ap:AAA} details our automatic atomic-action annotation procedure.

\textbf{Visual and tactile latent targets.}
Given action-word labels paired with future sensory observations, we construct visual latent targets following prior latent VLA methods~\cite{liu2026last}. At each target timestep, we encode the future head-view image using Qwen3-VL's visual encoder and pool its final-layer spatial tokens into a single latent target for the corresponding visual Physical Words.
All tactile observations are derived from visuotactile signals. We construct future tactile targets separately for each end-effector using the same T3 encoder and feature aggregation as the current tactile inputs. These targets supervise tactile Physical Words to anticipate future contact information. Future observations serve only as training targets and are not used at inference.

\subsection{Temporally Structured PhysWord Space}
\label{sec:physword_space}
Manipulation unfolds continuously as action semantics and contact conditions evolve. A single future Physical Word may miss intermediate transitions within an action chunk, such as approaching an object before picking it up. We therefore organize Physical Words over multiple future timesteps to provide action generation with temporally structured predictive context.

\textbf{Temporal organization.}
To reduce the inference latency of dense Physical Word prediction, we predict Physical Words every $s$ steps over the horizon $H$, yielding $K=H/s$ future prediction times: $t+s, t+2s, \ldots, t+H$, where $t$ is the current timestep. 
Let $\mathbf{z}_{t,k}^{r,m}$ denote the contextual output hidden state used to predict a Physical Word for future time $t+ks$, where $k\in\{1,\ldots,K\}$.
Here,  $r\in\{\mathrm{L},\mathrm{R}\}$ denotes the left or right end-effector, and $m\in\{\mathrm{v},\mathrm{t}\}$ denotes the visual or tactile Physical Word variant. We organize these representations as
\begin{equation}
    \mathbf{Z}_t =
    \begin{bmatrix}
        B_{t,1}^{\mathrm{L}} &
        B_{t,1}^{\mathrm{R}} &
        \cdots &
        B_{t,K}^{\mathrm{L}} &
        B_{t,K}^{\mathrm{R}}
    \end{bmatrix},
    \qquad
    B_{t,k}^{r}
    =
    \bigl(
        \mathbf{z}_{t,k}^{r,\mathrm{v}},
        \mathbf{z}_{t,k}^{r,\mathrm{t}}
    \bigr).
\label{eq:physword_temporal_layout}
\end{equation}
Blocks are ordered by future timestep, with the left end-effector preceding the right. Each block $B_{t,k}^{r}$ contains the visual and tactile predictions for end-effector $r$ at time $t+ks$.

\textbf{Temporal blockwise generation.}
As shown in Figure~\ref{fig:method}(c), PhysWord generates Physical Words autoregressively over future timestep blocks, jointly predicting visual and tactile variants for each end-effector. 
For $k=1$, we initialize generation with a Physical Word start token appended to the task instruction.
We apply a full-block input--target shift: for $k\geq2$, each block receives word embeddings from $t+(k-1)s$, while its output hidden states receive action-label and sensory-latent supervision for $t+ks$. Attention is full within each block and causal across blocks of the same end-effector, with cross-effector attention blocked. 
The complete attention mask is provided in Figure~\ref{fig:apattention_mask}.

\subsection{Training Objectives}
\label{sec:training_objectives}

We train PhysWord in two stages. In Stage 1, we pretrain the model on approximately 7,000 hours of human-hand and robot demonstrations with and without tactile observations. Details of the pretraining data and configuration are provided in Appendix~\ref{ap:pd}. In Stage~2, we fine-tune the PhysWord on task-specific demonstrations for downstream manipulation.
For each stage, we jointly optimize four objectives: the Cosmos flow-matching loss $L_{\mathrm{video}}$, Physical Word cross-entropy loss $L_{\mathrm{CE}}$, future latent prediction loss $L_{\mathrm{latent}}$, and action loss $L_{\mathrm{act}}$.
The latent alignment loss is defined as:
\begin{equation}
   L_{\mathrm{latent}}
    =
    1-\frac{\mathbf{z}_t^{\top}\mathbf{g}_t}
    {\|\mathbf{z}_t\|_2\|\mathbf{g}_t\|_2+\epsilon},
\end{equation}
where $\epsilon$ ensures numerical stability. For brevity, end-effector, future-offset, and modality indices are omitted in the loss notation. We average this term over all valid latent predictions to obtain $\mathcal{L}_{\mathrm{latent}}$.
The total objective combines all four losses:
$
    L
    =
    L_{\mathrm{video}}
    + L_{\mathrm{CE}}
    + L_{\mathrm{latent}}
    + L_{\mathrm{act}}.
$

\section{Experiments}
\label{sec:experiments}

We compare PhysWord with baselines on two simulation benchmarks (Section~\ref{sec:simulation}), assess component contributions through ablations (Section~\ref{sec:ablations}), and evaluate real-world performance and Physical Word reuse across unseen configurations and long-horizon tasks (Section~\ref{sec:real-world}).

\subsection{Simulation Experiments}
\label{sec:simulation}

\textbf{Benchmarks and Protocol.}
We evaluate PhysWord on two benchmarks: UniVTAC for contact-rich manipulation with visuotactile sensing and LIBERO-Long for long-horizon manipulation.
\textbf{UniVTAC}~\cite{chen2026univtac} comprises eight tasks spanning lifting, classification, and insertion.
\textbf{LIBERO-Long}~\cite{liu2023liberobenchmarkingknowledgetransfer} comprises ten long-horizon tasks. 
For both benchmarks, we use 50 training demonstrations per task and follow the official success criteria. We evaluate PhysWord over three runs of 100 rollouts per task and compute each task’s standard deviation across runs.

\textbf{Baselines and Implementation details.}
On UniVTAC, we compare PhysWord with two categories of baselines:
(1) \textit{vision-based models}, including $\pi_{0.5}$~\cite{intelligence2025pi05visionlanguageactionmodelopenworld}, StarVLA-$\alpha$~\cite{ye2026starvla}, and InternVLA-A1~\cite{cai2026internvla}, and
(2) \textit{tactile-enabled policies}, including VITaL~\cite{george2025vital}, ACT$^{Tac}$ with the UniVTAC Encoder, FTP-1~\cite{yuan2026ftp}, N0-VTLA~\cite{team2026n_0}, T-Rex~\cite{niu2026t}, and TacDyn-WAM~\cite{wang2026tacdyn}.
On LIBERO-Long, we compare PhysWord with two categories of baselines:
(1) \textit{VLA models}, including OpenVLA~\cite{kim2024openvla}, $\pi_{0.5}$, OpenVLA-OFT~\cite{kim2025finetuningvisionlanguageactionmodelsoptimizing}, GR00T-N1.7~\cite{bjorck2025gr00t}, InternVLA-A1, and LaST$_0$~\cite{liu2026last}, and
(2) \textit{WAMs}, including Motus~\cite{bi2025motus}, Fast-WAM~\cite{yuan2026fast}, and UniWAM~\cite{chen2026uniwam}.
Across both benchmarks, we report official baseline results or reproduce them through full fine-tuning following each method's official setup.
On UniVTAC, we train a separate policy for each task, following FTP-1.
On LIBERO-Long, we jointly train a single policy on all tasks.
Detailed implementation settings for PhysWord and the baselines are provided in Appendix~\ref{app:implementation_details}.

\begin{table*}[t]
\centering
\caption{
Comparison on UniVTAC (left) and LIBERO-Long (right) in terms of success rates (\%). Bold indicates the best performance for each task.
}
\label{tab:simulation-main}
\small
\setlength{\tabcolsep}{4pt}
\renewcommand{\arraystretch}{1.12}
\resizebox{\textwidth}{!}{
\begin{tabular}{@{}l*{9}{c}@{\hspace{1.5em}}lc@{}}
\hline
\multicolumn{10}{c}{\textbf{UniVTAC (Visuotactile simulator)}}
& \multicolumn{2}{c}{\textbf{LIBERO-Long}} \\
\cmidrule(lr){1-10}
\cmidrule(lr){11-12}
\textbf{Method}
& \textbf{\shortstack{Lift\\Bottle}}
& \textbf{\shortstack{Insert\\Tube}}
& \textbf{\shortstack{Insert\\Hole}}
& \textbf{\shortstack{Grasp\\Classify}}
& \textbf{\shortstack{Pull Out\\Key}}
& \textbf{\shortstack{Put\\Bottle}}
& \textbf{\shortstack{Insert\\HDMI}}
& \textbf{\shortstack{Lift\\Can}}
& \textbf{\shortstack{Mean\\S.R. $\uparrow$}}
& \textbf{Method}
& \textbf{S.R. $\uparrow$} \\
\midrule
ACT$^{Tac}$
& 71 & 56 & 24 & 99 & 46 & 31 & 28 & 29 & 48.0
& OpenVLA & 53.7  \\

VITaL
& 72 & 34 & 25 & \textbf{100} & 47 & 32 & 6 & 8 & 40.5
& $\pi_{0.5}$ & 92.4 \\

$\pi_{0.5}$
& \textbf{100} & 74 & 25 & 49 & 35 & 34 & 8 & 72 & 49.6
& OpenVLA-OFT & 94.5\\

StarVLA-$\alpha$
& 32 & 69 & 52 & 68 & 51 & 88 & 24 & 65 & 56.1
& GR00T-N1.7 & 94.4 \\

InternVLA-A1
& 58 & 95 & 93 & 86 & 66 & 37 & 12 & 90 & 67.1
& LaST$_0$ & 95.6\\

FTP-1
& 97 & 79 & 64 & 99 & 48 & 47 & 4 & 65 & 62.9
& Motus & 97.6 \\

N0-VTLA
& 99 & 99 & 95 & \textbf{100} & 99 & 60 & 25 & 88 & 83.1
& Fast-WAM & 95.2 \\ 

T-Rex
& 99 &  95&  \textbf{100}&  95&  99&  79&  8&  75& 81.3
&  InternVLA-A1 & 98.4  \\

TacDyn-WAM& 97 &  97&  91&  99&  97&  72&  12&  87& 81.5
&  UniWAM & 98.4  \\

\midrule
\textbf{PhysWord}
& \textbf{100} & \textbf{100} & \textbf{100} & \textbf{100}
& \textbf{100} & \textbf{93} & \textbf{34} & \textbf{95}
& \textbf{90.3}
& \textbf{PhysWord} & \textbf{99.6} \\
\hline
\end{tabular}
}
\end{table*}

\textbf{Results on UniVTAC.}
PhysWord achieves a mean success rate of 90.3\% (average per-task standard deviation (SD): 1.5 percentage points), surpassing the strongest tactile baseline, N0-VTLA, by 7.2 percentage points and the strongest vision-based VLA baseline, InternVLA-A1, by 23.2 percentage points. Gains are particularly pronounced on \textit{Put Bottle in Shelf} and \textit{Lift Can}, suggesting benefits for sustained motion coordination and contact-sensitive manipulation.
We further evaluate \textbf{PhysWord-feedback}, which replaces measured tactile inputs with predicted tactile latents.
It achieves 89\% mean success, only 1.3 percentage points below the full model, suggesting that the predicted tactile latents faithfully capture contact information needed for control. These results indicate that atomic action semantics can anchor future dynamics while preserving the fidelity of tactile predictions.
Per-task standard deviations for PhysWord are reported in Appendix~\ref{app:additional_simulation}, while detailed experiments on predictive tactile feedback in simulation and the real world are detailed in Appendix~\ref{app:real_tactile_feedback}.

\textbf{Results on LIBERO-Long.}
We also evaluate PhysWord on the vision-only benchmark, where it predicts only visual Physical Words. As shown in Table~\ref{tab:simulation-main}, PhysWord achieves a mean success rate of 99.6\%, with an average per-task SD of 0.3 percentage points. It outperforms $\pi_{0.5}$ (92.4\%), LaST$_0$ (95.6\%), and UniWAM (98.4\%) by 7.2, 4.0, and 1.2 percentage points, respectively.
These gains over VLA, latent reasoning, and WAM baselines suggest that temporally structured Physical Words provide effective predictive context for long-horizon action generation. By organizing future visual dynamics around atomic action semantics, they capture transitions between task stages.

\textbf{Model inference speed.}
With BF16 precision, five action denoising steps, and a Physical Word prediction interval of $s=4$, PhysWord runs at 4.9\,Hz on an NVIDIA H100 GPU. An optimized inference runtime increases throughput to 10.6\,Hz through kernel fusion, caching of reusable metadata, and parallel preprocessing and encoding. 
Details of these strategies are provided in Appendix~\ref{app:inference_efficiency}.

\subsection{Ablation Studies}
\label{sec:ablations}

We conduct ablation studies on LIBERO-Long and two UniVTAC tasks (\textit{Lift Can} and \textit{Put Bottle}) to examine Physical Word structure and temporal organization. Appendix~\ref{app:additional_ablations} presents further ablations on tactile inputs and prediction, visual target selection, and Physical Word embedding design. Appendix~\ref{app:additional_simulation} shows additional results for PhysWord without the temporal perception branch.

\begin{figure*}[t]
\centering
\begin{minipage}[c]{0.46\linewidth}
\centering
\small
\setlength{\tabcolsep}{4pt}
\renewcommand{\arraystretch}{1.15}

\begin{tabular*}{\linewidth}{
    @{\extracolsep{\fill}}ccccc@{}
}
\hline
& & \multicolumn{2}{c}{\textbf{UniVTAC}}
& \textbf{LIBERO} \\
\cmidrule(lr){3-4}
\cmidrule(l){5-5}
\textbf{Group}
& \textbf{Variant}
& \textbf{Bottle}
& \textbf{Can}
& \textbf{Avg.} \\
\hline

\multirow{4}{*}{1}
& $O$  &77 &83 & 97.0 \\
& $T$  &73 &65 & -- \\
& $W$  &77 & 67 & 97.4 \\

\hline
\multirow{2}{*}{2}
& $WO$ & 87& 89& 99.6\\
& $WT$ & 88& 75& -- \\

\hline
\multirow{2}{*}{3}
& $WT\!\rightarrow\!WO$ & 90& 90& -- \\
& $WO\!\rightarrow\!WT$ & 93& 95& -- \\

\hline
\end{tabular*}
\end{minipage}
\hfill
\begin{minipage}[c]{0.50\linewidth}
    \centering
    \includegraphics[width=0.91\linewidth]{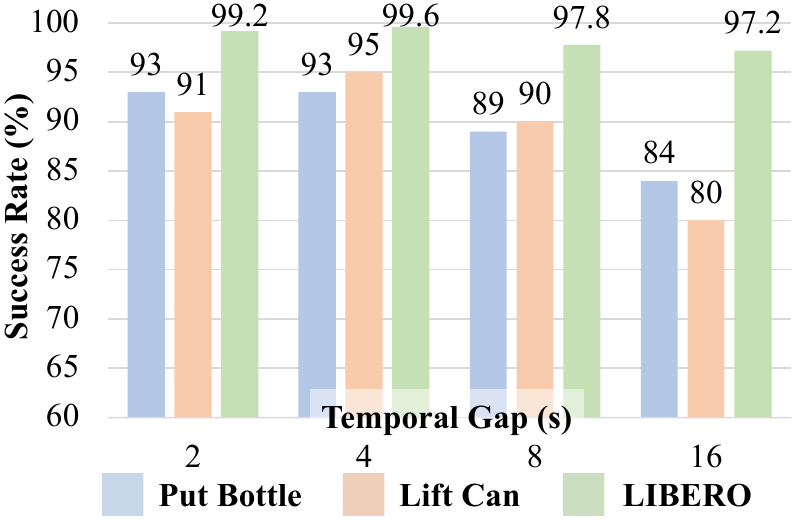}
\end{minipage}
\vspace{-0.1cm}
\caption{
\textbf{Ablation study.}
Left: Physical Word structure ablations on two UniVTAC tasks and LIBERO-Long.
Right: effect of the prediction interval in the temporally structured PhysWord space.
}
\label{fig:supervision-ablation}
\end{figure*}

\subsubsection{Effectiveness of Physical Word Latent Reasoning}
\label{sec:EPWLR}

We compare multimodal structures and supervision objectives to assess their contributions to Physical Word latent reasoning. Figure~\ref{fig:supervision-ablation} (left) summarizes three groups of ablations. Our PhysWord achieves the best overall performance on both visuotactile UniVTAC and vision-only LIBERO-Long. All variants with temporal prediction use the best-performing sampling interval of $s=4$.

\textbf{Group 1: Individual supervision signals.}
We first evaluate visual latent supervision alone ($O$), tactile latent supervision alone ($T$), and atomic-word prediction alone ($W$).
For the latent-only variants, we follow LaST$_0$ and autoregressively predict latent tokens using the same temporal organization as PhysWord.
These variants underperform joint semantic and latent supervision, suggesting that action semantics and future dynamics provide complementary guidance for latent reasoning.

\textbf{Group 2: Joint semantic and latent supervision.}
Combining atomic-word prediction with visual ($WO$) or tactile ($WT$) latent supervision improves performance over the corresponding single-objective variants in Group~1.
These gains suggest that shared action labels serve as semantic anchors for latent learning, helping the model capture common action semantics while retaining context-specific physical details.
We further evaluate $WO_{\mathrm{MLP}}$ (82\% on Put Bottle and 78\% on Lift Can) and $WT_{\mathrm{MLP}}$ (80\% on Put Bottle and 70\% on Lift Can), both of which apply a learnable MLP to the contextual hidden states before aligning them with future latent targets. Both variants perform worse than direct hidden-state supervision. These results suggest that directly supervising the word hidden states more effectively couples future physical dynamics with the action semantics.

\textbf{Group 3: Multimodal reasoning structure.}
With direct latent supervision, we compare two within-block token orders: tactile before visual ($WT\rightarrow WO$) and visual before tactile ($WO\rightarrow WT$). Both orders outperform the unimodal variants, suggesting that visual and tactile Physical Words provide complementary predictive context. 
We further compare pretrained VLM word embeddings with random initialization for both orderings. On \textit{Put Bottle} and \textit{Lift Can}, random initialization reduces success rates from 90\%/90\% to 84\%/86\% for $WT\rightarrow WO$, and from 93\%/95\% to 89\%/90\% for $WO\rightarrow WT$, respectively. These results highlight the value of inheriting word embeddings from large-scale VLM pretraining for learning semantically structured physical latent representations.

\textbf{Representation analysis.}
To examine the effect of atomic-word supervision beyond policy performance, we analyze intra-class latent distances using 50 examples for each of the four most frequent atomic actions in our downstream real-world data: \textit{pick}, \textit{place}, \textit{approach}, and \textit{move}.
We compare latent-only supervision ($O$, $T$) with Physical Words ($WO$, $WT$) using identical samples within each modality.
We compute the mean pairwise cosine distance between contextual hidden states within each action class.
Averaged across the four actions, word supervision reduces visual distances from $5.13\times10^{-3}$ ($O$) to $3.03\times10^{-3}$ ($WO$) and tactile distances from $2.50\times10^{-4}$ ($T$) to $1.25\times10^{-4}$ ($WT$), corresponding to reductions of approximately 41\% and 50\%, respectively.
Meanwhile, Figure~\ref{fig:vis}(b) compares latent-to-image attention maps for $O$/$W$/$WO$ and $T$/$WT$. 
As highlighted by the red boxes, joint word and latent supervision concentrates attention on manipulated objects and task-relevant regions. Together, these findings suggest that semantic anchoring promotes consistent action-related representations across contexts while reducing sensitivity to incidental scene variations. 
Appendix~\ref{app:representation_analysis} provides detailed intra-class distance results and t-SNE visualizations (Figure~\ref{fig:tsne}).

\subsubsection{Temporal Prediction Interval}
We evaluate Physical Word prediction intervals of $2$, $4$, $8$, and $16$ with a fixed action horizon of $H=16$. For example, an interval of $2$ yields predictions at $t+2,t+4,\ldots,t+16$.
As shown in Figure~\ref{fig:supervision-ablation}, shorter intervals generally improve performance, suggesting that denser predictions capture intermediate action transitions and changes in physical dynamics that coarser sampling may miss.
We select an interval of $4$ to balance manipulation success and inference efficiency, preserving informative intermediate predictions while limiting reasoning overhead.

\subsection{Real-World Experiments}
\label{sec:real-world}

\begin{figure}[t]
    \centering
    \includegraphics[width=\textwidth]{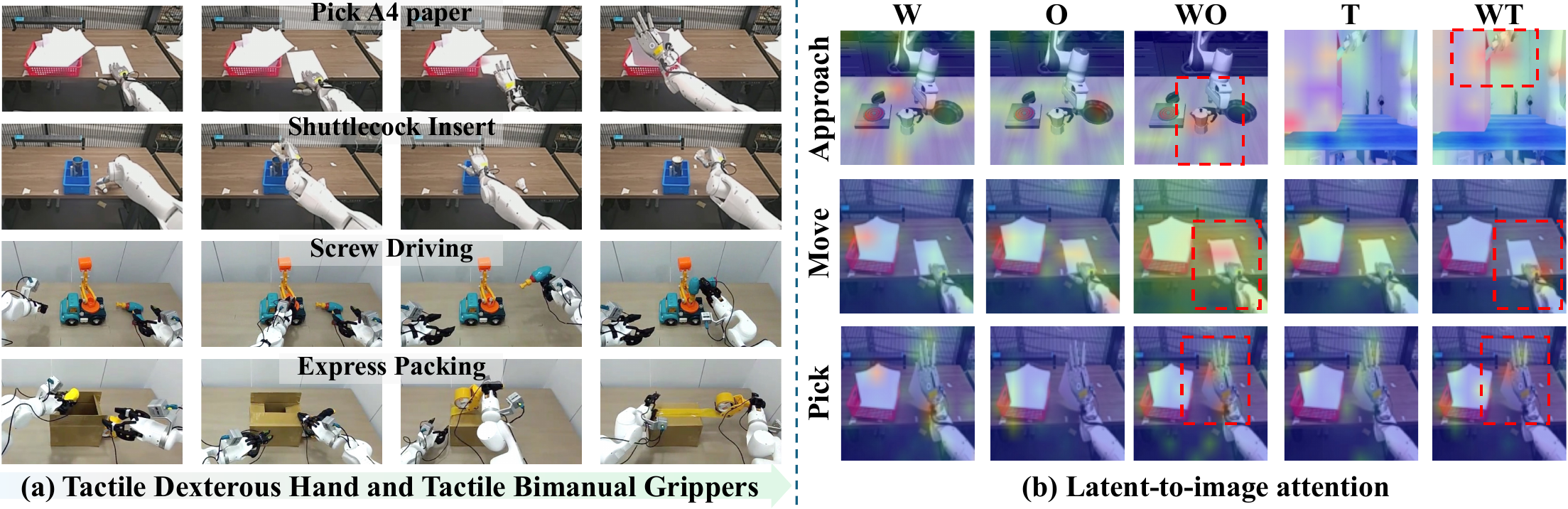} 
    \vspace{-0.5cm}
 \caption{\textbf{Qualitative results.}
(a) Real-world task execution with PhysWord.
(b) Latent-to-image attention maps across simulated and real-world tasks.
}
    \label{fig:vis}
\end{figure}

\textbf{Real-world experimental setup.}
We evaluate PhysWord on two robotic platforms equipped with dexterous hands and parallel grippers, respectively.
For dexterous manipulation, we use Tianji arms equipped with Sharpa Wave hands featuring tactile sensing on all fingers. As shown in Figure~\ref{fig:vis} (a), we evaluate contact-rich \textit{Pick Paper} and long-horizon \textit{Shuttlecock Insertion}.
For gripper-based manipulation, we use an i7 robot with Xense tactile sensors mounted on its gripper fingertips, evaluating long-horizon \textit{Express Packing} and precision-demanding \textit{Powered Screw Driving}. Additional details of the robotic platforms and data collection are provided in Appendix~\ref{ap:res}.

\textbf{Baselines and implementation details.}
We compare PhysWord with $\pi_{0.5}$, a representative VLA model, and FTP-1, a recent tactile foundation policy.
All methods receive the same head- and wrist-view RGB observations. FTP-1 and PhysWord additionally receive tactile observations.
For each task, we collect 100 training demonstrations and evaluate each method over three runs of 30 trials each.
Additional details of the evaluation protocol are provided in Appendix~\ref{app:evaluation_protocol}.

\begin{table*}[t]
\centering
\caption{
\textbf{Real-world evaluation.}
Success rates (\%) on four real-world manipulation tasks and on Pick Paper and Screw Drive with unseen backgrounds (Bac), objects (Obj), and positions (Pos).
Drop denotes the relative change from the in-domain mean for the same two tasks.
}
 \vspace{-0.2cm}
\label{tab:real-world}

\begin{minipage}[t]{0.47\linewidth}
\vspace{0pt}
\centering
\textbf{(a) Manipulation performance}\\[1mm]
\small
\setlength{\tabcolsep}{3pt}
\renewcommand{\arraystretch}{1.15}

\resizebox{\linewidth}{1.35cm}{%
\begin{tabular}{@{}lccccc@{}}
\toprule
\textbf{Method}
& \textbf{\shortstack{Pick\\Paper}}
& \textbf{\shortstack{Shuttlecock\\Insert}}
& \textbf{\shortstack{Express\\Packing}}
& \textbf{\shortstack{Screw\\Drive}}
& \textbf{\shortstack{Mean\\$\pm$ Avg. SD}} \\
\midrule
$\pi_{0.5}$ & 53.3 & 43.3 & 46.7 & 40.0 & 45.8$\pm$3.6 \\
FTP-1 & 66.7 & 56.7 & 50.0 & 46.7 & 55.0$\pm$4.6 \\
\textbf{PhysWord} & 93.3 & 60.0 & 73.3 & 60.0 & 71.7$\pm$2.5 \\
\bottomrule
\end{tabular}%
}
\end{minipage}
\hfill
\begin{minipage}[t]{0.51\linewidth}
\vspace{0pt}
\centering
\textbf{(b) Unseen configurations}\\[1mm]
\small
\setlength{\tabcolsep}{3pt}
\renewcommand{\arraystretch}{1.15}

\resizebox{\linewidth}{1.35cm}{%
\begin{tabular}{@{}lcccccccc@{}}
\hline
\textbf{Method}
& \multicolumn{3}{c}{\textbf{Pick Paper}}
& \multicolumn{3}{c}{\textbf{Screw Drive}}
& \textbf{Mean}
& \multirow{2}{*}{\textbf{Drop}} \\
\cmidrule(lr){2-4}
\cmidrule(lr){5-7}
Unseen:
& Bac & Obj & Pos
& Bac & Obj & Pos
& $\pm$ Avg. SD. & \\
\hline
$\pi_{0.5}$
& 46.7 & 40.0 & 13.3 & 26.7 & 20.0 & 26.7
& $28.9 \pm 3.8$ & -38\% \\
FTP-1
& 60.0 & 46.7 & 26.7 & 23.3 & 23.3 & 30.0
& $35.0 \pm 6.2$ & -38\% \\
\textbf{PhysWord}
& 90.0 & 83.3 & 80.0 & 43.3 & 36.7 & 46.7
& $63.3 \pm 3.3$ & -17\% \\
\bottomrule
\end{tabular}%
}
\end{minipage}
\end{table*}

\textbf{Real-world results.}
Table~\ref{tab:real-world}(a) shows that PhysWord outperforms both baselines on all four tasks, achieving a mean success rate of 71.7\%, compared with 45.8\% for $\pi_{0.5}$ and 55.0\% for FTP-1.
Improvements on \textit{Pick Paper} and \textit{Shuttlecock Insertion} demonstrate its effectiveness for dexterous manipulation requiring fine contact control. Gains on \textit{Express Packing} and \textit{Powered Screw Driving} further support its applicability to long-horizon manipulation and contact-sensitive tool use.

\textbf{Unseen configurations.}
We directly evaluate all methods on unseen backgrounds, objects, and positions, ensuring that none of the test configurations appear in the training data.
As shown in Table~\ref{tab:real-world}(b), PhysWord achieves 63.3\% mean success, outperforming $\pi_{0.5}$ and FTP-1 by 34.4 and 28.3 percentage points, respectively.
We evaluate Physical Word predictions against GPT annotations on 2,000 frames sampled across all rollouts and configurations, covering \textit{approach}, \textit{move}, \textit{pick}, and \textit{place}, and obtain 90.5\% frame-level accuracy.
Together, these results suggest that Physical Words preserve action semantics across contexts, providing reusable semantic anchors for predicting physical dynamics and supporting robust manipulation.

\textbf{Long-horizon extension.}
We further evaluate extended versions of \textit{Pick A4 Paper} and \textit{Express Packing} involving task sequences absent from training, requiring repeated pick-and-place cycles and revisiting packing stages. PhysWord outperforms both baselines by 70.0 percentage points on \textit{Pick A4 Paper}, and exceeds $\pi_{0.5}$ and FTP-1 by 40.0 and 36.7 percentage points, respectively, on \textit{Express Packing}. During repeated paper manipulation, the predicted Physical Word switches from \texttt{place} back to \texttt{approach} when a new sheet is introduced, providing context for the next cycle. These results suggest that Physical Words support the reuse of learned action primitives as task conditions evolve, enabling execution beyond the sequences observed during training. Appendix~\ref{app:long_horizon} provides detailed experimental setups and per-task results, along with additional experiments examining the contribution of Physical Words to executing extended long-horizon sequences.

\section{Conclusion and Limitation}
In this paper, we presented PhysWord, a reasoning-before-acting foundation model that uses atomic action semantics to structure latent representations of future physical dynamics. Its temporally organized visual and tactile Physical Words capture evolving motion and contact dynamics, providing predictive context for action generation. This temporal structure also supports the reuse of action primitives across task stages. Experiments on simulation benchmarks and real-world platforms demonstrate strong manipulation performance, generalization to unseen configurations, and effective execution of long-horizon tasks. Our current action vocabulary does not cover whole-body control, such as climbing stairs, which we leave for future work.

\bibliographystyle{unsrt}
\bibliography{references}  

\clearpage

\appendix

\begin{itemize}
    \item Section~\ref{ap:pd}: Pretraining Data and Details
    \item Section~\ref{ap:res}: Real-World Experimental Setup
    \item Section~\ref{app:additional_quantitative_results}: Additional Quantitative Results
    \item Section~\ref{app:representation_analysis}: Additional Representation Analysis
    \item Section~\ref{ap:fca}: Failure Case Analysis
    \item Section~\ref{app:implementation_details}: Additional Implementation Details
    \item Section~\ref{app:evaluation_protocol}: Real-World Evaluation Protocol
\end{itemize}

\section{Pretraining Data and Details}
\label{ap:pd}

\subsection{Data Sources and Composition}
We construct the pretraining corpus using 2,000 hours of human demonstrations from Egoscale~\cite{zheng2026egoscale}, the full public releases of DROID~\cite{khazatsky2024droid} and OpenNeoData~\cite{team2026n_0}, and real-world Sharpa demonstrations from T-Rex~\cite{niu2026t}. These sources cover human-hand manipulation and robot manipulation with and without tactile observations across different end-effectors. Table~\ref{tab:pretraining-data} summarizes the data composition.
Each training example pairs observation context with action targets, atomic action labels, and future sensory targets according to the available modalities. Future observations serve only as training targets and are not provided as policy inputs at inference.
Note that our pretraining focuses on establishing Physical Word reasoning capabilities. We deliberately use a smaller manipulation corpus than those used by our main baselines, e.g., $\pi_{0.5}$, N0-VTLA, and T-Rex, to evaluate our approach without an advantage in pretraining data volume.

\begin{table}[!ht]
\centering
\caption{Pretraining data composition and volume.}
\label{tab:pretraining-data}
\small
\begin{tabular*}{\linewidth}{@{\extracolsep{\fill}}lrc@{}}
\toprule
\textbf{Dataset} & \textbf{Duration (h)} & \textbf{Tactile} \\
\midrule
Human hand data & 2,000+ & No \\
DROID & 350 & No \\
OpenNeoData & 5,000+ & Yes \\
Sharpa data & 100+ & Yes \\
\midrule
Total & 7,400+ & -- \\
\bottomrule
\end{tabular*}
\end{table}

\subsection{Data Preprocessing}
\textbf{Visual and action quality filtering.}
Visual filtering identifies black or near-uniform frames, severe underexposure or overexposure, motion blur, and prolonged periods when the manipulating hand or gripper and relevant objects are not visible in any available view. For action quality, we discard records with invalid values or malformed entries and detect abnormal translation or rotation jumps, unusually short or long trajectories, frozen motion, and excessive proportions of stationary steps. Filtering thresholds are calibrated using training data only. We group trajectories by embodiment, action representation, and control frequency, use the 1st and 99th percentiles ($Q_{01}$ and $Q_{99}$) as initial references, and manually adjust the thresholds for each filtering criterion.

\textbf{Trajectory diversity filtering.}
To reduce redundant demonstrations, we compare trajectories within each dataset that share the same task and scene. We extract keyframes following the keyframe-discovery procedure of PerAct~\cite{shridhar2023perceiver} and compare wrist positions in the world coordinate frame at corresponding keyframes. Trajectories whose corresponding wrist positions differ by less than 2\,cm are treated as near-duplicates. We retain at most two demonstrations per group and discard the remainder to reduce motion redundancy that may limit spatial generalization.

\textbf{Action representations and retargeting.}
Robot demonstrations retain their original action spaces. For human-hand data, we follow EgoScale's processing protocol~\cite{zheng2026egoscale}, retaining wrist poses and retargeting finger motions to a 22-DoF dexterous hand. This retains the demonstrated wrist motion while providing hand-action targets compatible with the target hand kinematics.

\subsection{Atomic Action Annotation}
\label{ap:AAA}

\textbf{Annotation method.} We use GPT~\cite{openai2026gpt56} to segment processed demonstrations into atomic action stages and assign labels from our action vocabulary. For bimanual demonstrations, the left and right end-effectors are annotated independently, allowing simultaneous actions to receive distinct labels. We record each stage's temporal boundaries and align its label with the corresponding observations, providing cross-entropy targets paired with future sensory supervision. Visual and tactile Physical Words use modality-specific variants of the same action label, with tactile variants and their latent supervision included only when tactile observations are available.

\textbf{Annotation quality assessment.}
We randomly sample 500 examples from the pretraining corpus and manually annotate their atomic actions. GPT-generated labels achieve 95\% accuracy against these human annotations. Errors primarily involve ambiguous or incidental movements. For example, finger tapping on a table during pauses in manipulation may be mislabeled as \textit{click}, despite the absence of a button press. These cases highlight the difficulty of distinguishing intentional manipulation from incidental motion. Nevertheless, VLM-based video annotation provides a practical, scalable pipeline for fine-grained atomic-action labeling. PhysWord maintains stable task execution despite these occasional ambiguities, suggesting robustness to limited annotation noise.

\subsection{Pretraining Configuration}
We pretrain PhysWord for two epochs using the same optimization configuration as downstream fine-tuning (Appendix~\ref{app:implementation_details}). We use AdamW with zero weight decay and a per-GPU batch size of 6. The base learning rate is $10^{-4}$, with separate rates of $2\times10^{-6}$ for the Cosmos backbone and $5\times10^{-5}$ for the T3 tactile encoder. 
Training uses BF16 mixed precision and linear warmup over the first 5\% of optimizer steps, followed by cosine learning-rate decay to zero.

\section{Real-World Experimental Setup}
\label{ap:res}

\begin{figure}[h]
    \centering
    \includegraphics[width=0.9\linewidth]{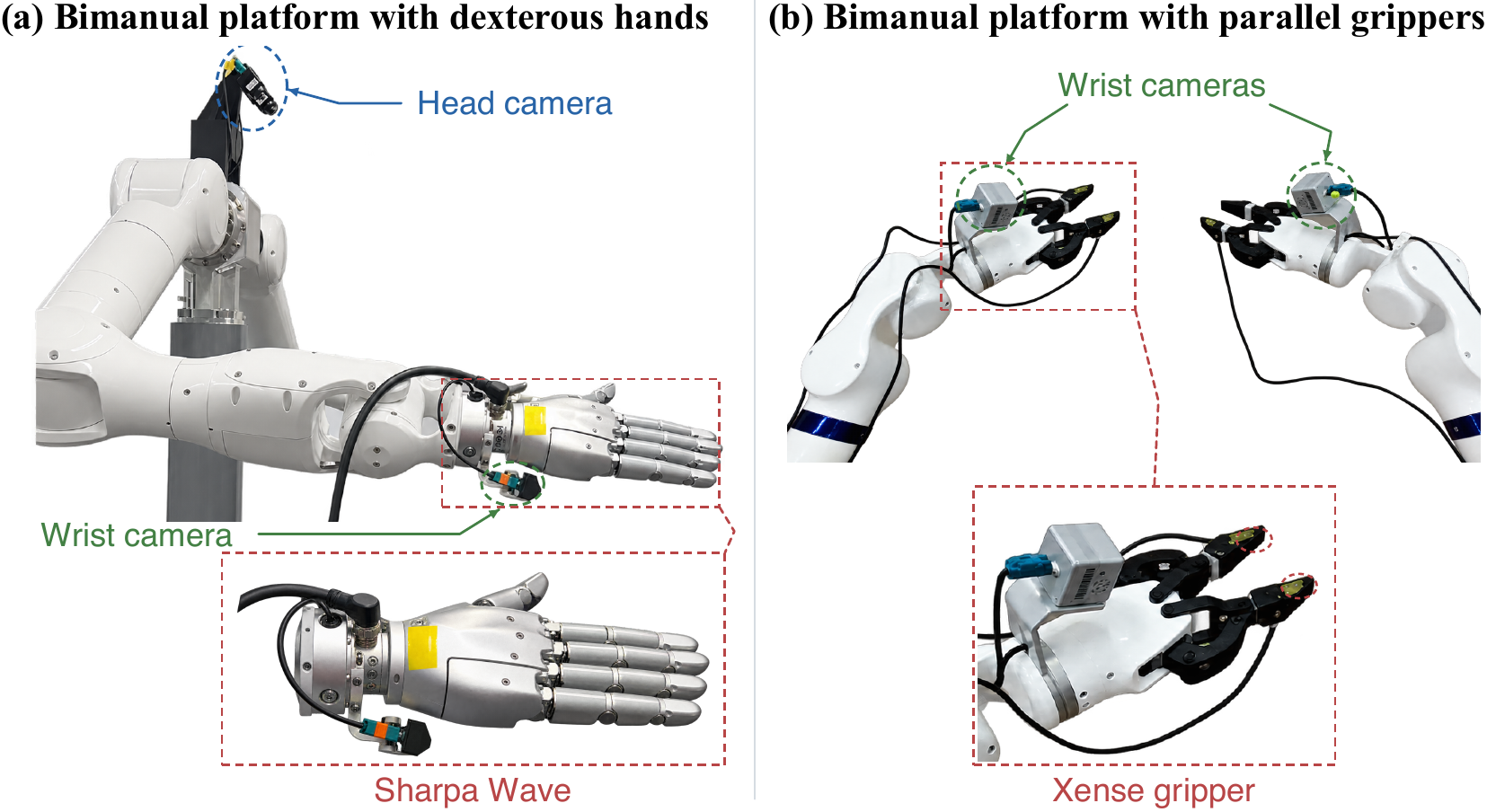}
    \caption{\textbf{Real-world robotic platforms and sensing.}
    Representative hardware views of (a) the bimanual platform with
    Sharpa Wave dexterous hands and (b) the bimanual platform with
    Xense tactile grippers. Blue and green annotations identify head and
    wrist cameras, respectively; red insets highlight the end effectors.
    }
    \label{fig:real-world-setup}
\end{figure}

\subsection{Robotic Platforms and Sensing}
Figure~\ref{fig:real-world-setup} presents the two bimanual platforms used in our experiments. The dexterous-hand platform comprises Tianji Marvin arms equipped with two Sharpa Wave hands, each featuring 22 active degrees of freedom and five-finger tactile sensing. The gripper platform uses an i7 Pro robot with Xense tactile sensors mounted on its parallel-gripper fingertips. Each arm has seven revolute joints, denoted J1--J7. Table~\ref{tab:arm-joint-limits} reports their joint angle limits, excluding end-effector joints. Together, these platforms support evaluation across gripper-based and high-dimensional dexterous manipulation. 
Both platforms provide three RGB views: one head view capturing the overall scene and two wrist views providing close-range observations of manipulation. Visuotactile observations are acquired from the gripper fingertips or dexterous-hand fingers and supplied to the policy alongside RGB observations and language instructions.

\begin{table}[t]
\centering
\caption{
Arm joint position limits in radians, excluding end-effector joints.
}
\label{tab:arm-joint-limits}
\small
\begin{tabular*}{\linewidth}{@{\extracolsep{\fill}}lccc@{}}
\toprule
Joint & i7 Pro left & i7 Pro right & Tianji \\
\midrule
J1 & $[-3.14,\,3.14]$ & $[-3.14,\,3.14]$ & $[-3.1067,\,3.1067]$ \\
J2 & $[-0.6977,\,3.8377]$ & $[-3.8377,\,0.6977]$ & $[-2.0944,\,2.0944]$ \\
J3 & $[-3.14,\,3.14]$ & $[-3.14,\,3.14]$ & $[-3.1067,\,3.1067]$ \\
J4 & $[-2.4248,\,2.4248]$ & $[-2.4248,\,2.4248]$ & $[-2.5307,\,1.0472]$ \\
J5 & $[-3.14,\,3.14]$ & $[-3.14,\,3.14]$ & $[-3.1067,\,3.1067]$ \\
J6 & $[-1.0467,\,1.0467]$ & $[-1.0467,\,1.0467]$ & $[-1.0472,\,1.0472]$ \\
J7 & $[-1.6572,\,1.8317]$ & $[-1.6572,\,1.8317]$ & $[-1.5708,\,1.5708]$ \\
\bottomrule
\end{tabular*}
\end{table}

\subsection{Downstream Task Data Collection}
We collect demonstrations through human teleoperation on both platforms. The bimanual gripper platform uses a PICO-based VR teleoperation interface. The dexterous-hand platform uses the same interface, augmented with Manus gloves to capture finger motions for dexterous-hand control. During teleoperation, we record multi-view RGB observations, tactile signals, robot proprioception, and action commands. We collect 100 demonstrations per task for downstream fine-tuning.

\section{Additional Quantitative Results}
\label{app:additional_quantitative_results}

We provide additional simulation results, controlled ablations, and
real-world evaluations to further examine the contributions of Physical
Words, temporal composition, and predictive tactile feedback.

\subsection{Additional Simulation Experiments}
\label{app:additional_simulation}

Table~\ref{tab:univtac-additional} reports per-task results on all eight
UniVTAC tasks, extending the main-paper comparison with PhysWord-Lite.

\textbf{PhysWord-Lite.}
To examine whether Physical Words remain effective without the temporal
perception branch, we remove Cosmos and supply all camera views to the
reasoning-and-action branch. Physical Word generation and its semantic
and latent supervision are retained.
PhysWord-Lite maintains strong performance on contact-rich manipulation
tasks, indicating that effective contact-sensitive control does not
depend solely on the temporal perception branch. These results highlight
the contribution of the physical latents encoded by Physical Words,
which provide action generation with semantically structured predictions
of future visual dynamics and contact states. The temporal branch
further strengthens perception, while Physical Words retain their
effectiveness even without this additional support.

\subsection{Additional Ablation Studies}
\label{app:additional_ablations}

We conduct three ablations on \textit{Put Bottle in Shelf} and
\textit{Lift Can}. Unless explicitly varied, we retain the full model's
training data, optimization settings, action objective, and evaluation
protocol, with a Physical Word prediction interval of $s=4$.

\textbf{Current tactile input and future tactile prediction.}
We distinguish the contribution of measured contact context from that of predictive tactile reasoning. The \textit{w/o tactile input} variant removes measured tactile features from both branches while retaining tactile Physical Words and their supervision. The \textit{w/o tactile prediction} variant removes tactile Physical Words and their associated supervision while retaining measured tactile inputs. The \textit{w/o both} variant removes both components, yielding a vision-only configuration. As shown in Table~\ref{tab:tactile_input_output}, all three variants underperform the full PhysWord model. Current tactile observations inform the policy about the existing contact state, whereas tactile Physical Words anticipate how contact will evolve during execution.

\begin{table*}[t]
\centering
\caption{
Additional results on UniVTAC, reporting task success rates (\%).
PhysWord-Lite removes the temporal perception branch.
PhysWord-feedback replaces online tactile measurements with predicted tactile latents during evaluation. The SD row is the average of the per-task SDs.
}
\label{tab:univtac-additional}
\small
\setlength{\tabcolsep}{4pt}
\renewcommand{\arraystretch}{1.12}
\resizebox{\textwidth}{!}{%
\begin{tabular}{@{}l*{9}{c}@{}}
\toprule
\textbf{Method}
& \textbf{\shortstack{Lift\\Bottle}}
& \textbf{\shortstack{Insert\\Tube}}
& \textbf{\shortstack{Insert\\Hole}}
& \textbf{\shortstack{Grasp\\Classify}}
& \textbf{\shortstack{Pull Out\\Key}}
& \textbf{\shortstack{Put\\Bottle}}
& \textbf{\shortstack{Insert\\HDMI}}
& \textbf{\shortstack{Lift\\Can}}
& \textbf{\shortstack{Mean\\S.R. $\uparrow$}} \\
\midrule
ACT$^{Tac}$
& 71 & 56 & 24 & 99 & 46 & 31 & 28 & 29 & 48.0 \\
VITaL
& 72 & 34 & 25 & 100 & 47 & 32 & 6 & 8 & 40.5 \\
$\pi_{0.5}$
& 100 & 74 & 25 & 49 & 35 & 34 & 8 & 72 & 49.6 \\
StarVLA-$\alpha$
& 32 & 69 & 52 & 68 & 51 & 88 & 24 & 65 & 56.1 \\
InternVLA-A1
& 58 & 95 & 93 & 86 & 66 & 37 & 12 & 90 & 67.1 \\
FTP-1
& 97 & 79 & 64 & 99 & 48 & 47 & 4 & 65 & 62.9 \\
N0-VTLA
& 99 & 99 & 95 & 100 & 99 & 60 & 25 & 88 & 83.1 \\
T-Rex
& 99 &  95&  100&  95&  99&  79&  8&  75& 81.3 \\
\midrule
\textbf{PhysWord}
& 100 & 100 & 100 & 100 & 100 & 93 & 34 & 95 & 90.3 \\
\textbf{(SD)}
 & $\pm 0$ & $\pm 0.5$ & $\pm 0$ & $\pm 0$ & $\pm 0.5$ & $\pm 2$  & $\pm 7$ & $\pm 2$ & $\pm 1.5$ \\ 
\midrule
PhysWord-Lite
& 99 & 99 & 100 & 98 & 100 & 83 & 29 & 83 & 86.4 \\
PhysWord-feedback
& 99& 99& 100& 98& 93& 92& 37& 94& 89 \\
\bottomrule
\end{tabular}%
}
\end{table*}

\begin{table}[t]
\centering
\caption{
Ablation of tactile input and prediction, reporting task success rates (\%).
}
\label{tab:tactile_input_output}
\small
\setlength{\tabcolsep}{6pt}
\renewcommand{\arraystretch}{1.12}
\begin{tabular}{@{}lcc@{}}
\toprule
\textbf{Variant}
& \textbf{Put Bottle}
& \textbf{Lift Can} \\
\midrule
Full PhysWord             & 93 & 95 \\
w/o tactile input         & 88 & 87 \\
w/o tactile prediction    &  87   &  89  \\
w/o both                  &  82 &  81  \\
\bottomrule
\end{tabular}
\end{table}

\textbf{View selection for visual latent targets.}
We compare visual latent supervision from head-view images, wrist-view images, and both views, while keeping the policy's observation inputs unchanged. When using both views, we generate an additional visual Physical Word at each sampled future timestep, with the two visual tokens supervised by their respective view-specific targets.
Table~\ref{tab:visual_target_views} shows that head-view targets yield the best performance. The head view captures object motion and spatial relations across the workspace, providing broader context for predicting future dynamics. Wrist views provide local detail but may become heavily occluded during contact with objects, limiting their predictive information.
These results support our use of head-view features as visual latent targets.

\begin{table}[t]
\centering
\caption{
Effect of visual latent target views on success rates (\%), with policy inputs unchanged.
}
\label{tab:visual_target_views}
\small
\setlength{\tabcolsep}{6pt}
\renewcommand{\arraystretch}{1.12}
\begin{tabular}{@{}lcc@{}}
\toprule
\textbf{Target view}
& \textbf{Put Bottle}
& \textbf{Lift Can} \\
\midrule
Head view             & 93 & 95 \\
Wrist views           & 85 & 83 \\
Head and wrist views  &  84  &  89  \\
\bottomrule
\end{tabular}
\end{table}

\begin{table}[t]
\centering
\caption{
Shared versus modality-specific Physical Words.
Entries report task success rates (\%).
}
\label{tab:physical_word_variants}
\small
\setlength{\tabcolsep}{6pt}
\renewcommand{\arraystretch}{1.12}
\begin{tabular}{@{}lcc@{}}
\toprule
\textbf{Physical Word design}
& \textbf{Put Bottle}
& \textbf{Lift Can} \\
\midrule
Shared: \texttt{pick}, \texttt{pick} & 90 & 89 \\
Separate: \texttt{pick\_v}, \texttt{pick\_t} & 93 & 95 \\
\bottomrule
\end{tabular}
\end{table}

\textbf{Shared versus modality-specific Physical Word embeddings.}
We compare repeating the same action token twice, e.g., \texttt{pick}, \texttt{pick}, with using separate modality-specific tokens, e.g., \texttt{pick\_v}, \texttt{pick\_t}. Both designs produce two contextual hidden states, supervised by visual and tactile latent targets, respectively. The repeated-token design shares one vocabulary embedding, whereas the modality-specific design uses independently trainable embeddings initialized from the same pretrained action word.
Modality-specific tokens achieve better performance
(Table~\ref{tab:physical_word_variants}). Although both modalities describe the same action, their targets occupy distinct representational spaces. This result suggests that separate vocabulary embeddings facilitate modality-specific adaptation while preserving a common semantic initialization, helping the corresponding hidden states capture complementary motion and contact information.

\subsection{Additional Real-World Experiments}
\label{app:additional_real_world}

\subsubsection{Long-Horizon Execution beyond Training Trajectories}
\label{app:long_horizon}

We investigate long-horizon generalization on extended versions of two real-world tasks, \textit{Pick A4 Paper} and \textit{Express Packing}. We also compare Physical Words with latent-only reasoning to assess how semantic anchoring supports the reuse of learned behaviors in task sequences absent from training.

\textbf{Experimental setup.}
We evaluate long-horizon generalization using extended task sequences absent from the training demonstrations, testing whether learned physical words can be reused as task conditions evolve.
For \textit{Pick A4 Paper}, training demonstrations contain a single pick-and-place cycle. During evaluation, a new sheet is introduced after each completed placement, requiring three consecutive cycles within one episode.
For \textit{Express Packing}, training demonstrations involve packing one object. At test time, the robot must pack two objects and repeat the closure stage after the box flaps are manually reopened, before sealing the box.
Success requires completing the entire extended sequence within a single episode, with every required stage satisfying the task-specific completion criteria in Appendix~\ref{app:evaluation_protocol}.
To complement the ablations in Section~\ref{sec:EPWLR}, we further examine the contribution of semantic anchoring on these extended tasks. For PhysWord, \textit{Extended} uses both visual and tactile Physical Words. The \textit{w/o Physical Words} variant retains visual and tactile latent supervision but removes atomic-word supervision, directly feeding predicted latents back as subsequent reasoning inputs instead of word embeddings.

\textbf{Results.}
As shown in the \textit{Extended} columns of Table~\ref{tab:long_horizon_generalization}, both baselines fail to complete extended \textit{Pick A4 Paper}, whereas PhysWord achieves 70.0\% success, demonstrating its ability to extend a learned single-cycle behavior into repeated execution.
On \textit{Express Packing}, whose training demonstrations already contain multiple stages, all methods exhibit some ability to handle the extension. PhysWord nevertheless achieves the highest success rate of 66.7\%, compared with 26.7\% for $\pi_{0.5}$ and 30.0\% for FTP-1.
Replacing Physical Words with latent-only reasoning reduces success to 36.7\% on extended \textit{Pick A4 Paper} and 50.0\% on extended \textit{Express Packing}, corresponding to decreases of 33.3 and 16.7 percentage points. The decline is particularly pronounced on \textit{Pick A4 Paper}, where training demonstrations contain only a single pick-and-place cycle.
Qualitative observations reveal that many failures occur at newly introduced stage transitions. After placing a sheet in the basket, PhysWord switches from \texttt{place} to \texttt{approach} when another sheet is introduced, providing explicit semantic context for the next cycle. The latent-only variant may instead remain near the previous placement location, exhibiting vertical oscillations without initiating the next reach. In these cases, another cycle is triggered only when the new sheet is placed at a specific location, suggesting greater dependence on spatial cues to resume execution.
These results suggest that semantic anchoring provides action-relevant guidance at stage transitions, supporting both repeated execution of single-cycle behaviors and adaptation of multi-stage behaviors to sequences absent from training.

\begin{table}[t]
\centering
\caption{Success rates (\%) on the original tasks and their extended long-horizon variants. For PhysWord, the Original and Extended columns both report results for the full model.}
\label{tab:long_horizon_generalization}
\small
\setlength{\tabcolsep}{4pt}
\renewcommand{\arraystretch}{1.15}
\begin{tabular}{@{}lcccccc@{}}
\toprule
\multirow{2}{*}{\textbf{Method}}
& \multicolumn{3}{c}{\textbf{Pick A4 Paper}}
& \multicolumn{3}{c}{\textbf{Express Packing}} \\
\cmidrule(lr){2-4}
\cmidrule(lr){5-7}
& Original & Extended & Extended (w/o PW)
& Original & Extended & Extended (w/o PW) \\
\midrule
$\pi_{0.5}$
& 53.3 & 0.0 &--
& 46.7 & 26.7 &-- \\
FTP-1
& 66.7 & 0.0 &--
& 50.0 & 30.0 &-- \\
\textbf{PhysWord}
& 93.3 & 70.0 & 36.7
& 73.3 & 66.7 &  50.0\\
\bottomrule
\end{tabular}
\end{table}

\subsection{Predictive Tactile Feedback}
\label{app:real_tactile_feedback}

\textbf{Simulation evaluation.}
We evaluate whether a policy trained with tactile observations can operate without online tactile sensing. Using the same trained PhysWord checkpoint, we replace measured tactile inputs with previously predicted tactile latents. For initial timesteps not yet covered by predictions, we initialize tactile context using observations from the opening frames of training demonstrations for the corresponding task.
In our simulation setup, only the first action of each predicted chunk is executed before replanning, yielding overlapping predictions for future timesteps. We apply temporal ensembling~\cite{zhao2023learning} to aggregate tactile latent estimates associated with the same target timestep and feed the resulting estimate back as tactile context when that timestep is reached.
As shown in Table~\ref{tab:univtac-additional}, PhysWord-feedback achieves performance comparable to PhysWord with measured tactile inputs. These results suggest that semantically anchored tactile latents retain the contact information needed for control, supporting deployment when tactile sensors are unavailable or fail.

\textbf{Real-world evaluation.}
We evaluate predictive tactile feedback on \textit{Pick A4 Paper} and \textit{Shuttlecock Insertion}, both of which require precise contact control. Paper manipulation relies on sustained frictional contact before grasping, whereas shuttlecock insertion requires a stable grasp and controlled contact during insertion and pressing.
Using the same trained checkpoint, we replace measured tactile inputs with predicted latents, following the initialization and temporal-ensembling procedure used in simulation. On real-world platforms, we execute the first eight actions of each predicted chunk, consistent with the main real-world experiments.
As shown in Table~\ref{tab:real_tactile_feedback}, PhysWord achieves success rates of 90.0\% and 56.7\% on Pick A4 Paper and Shuttlecock Insertion without online tactile sensing, compared with 93.3\% and 60.0\% using measured tactile inputs. The modest decrease of 3.3 percentage points on each task suggests that predicted latents preserve contact information needed for dexterous control. 
Together with the simulation results, these findings support deploying tactile-trained policies through predictive feedback when tactile measurements are unavailable.
We further quantify prediction accuracy by comparing all stored tactile latent predictions with their corresponding future ground-truth latents. These targets are extracted after each rollout from recorded tactile observations and matched to predictions by target timestep. The tactile recordings are used only for offline evaluation and are not supplied to the policy during these feedback rollouts. Across all predictions, the mean cosine similarity is 0.9873 and the mean squared error is 0.0630.

\begin{table}[t]
\centering
\caption{
Predictive tactile feedback on real-world dexterous tasks.
Entries report success rates (\%).
The two PhysWord variants use the same trained checkpoint.
}
\label{tab:real_tactile_feedback}
\small
\setlength{\tabcolsep}{5pt}
\renewcommand{\arraystretch}{1.15}
\begin{tabular}{@{}lccc@{}}
\toprule
\textbf{Method}
& \textbf{\shortstack{Online tactile\\sensing}}
& \textbf{\shortstack{Pick A4\\Paper}}
& \textbf{\shortstack{Shuttlecock\\Insertion}} \\
\midrule
$\pi_{0.5}$                  & No  & 53.3 & 43.3 \\
FTP-1                       & Yes & 66.7 & 56.7 \\
PhysWord                    & Yes & 93.3 & 60.0 \\
PhysWord w/ tactile feedback & No  & 90.0 & 56.7 \\
\bottomrule
\end{tabular}%
\end{table}

\begin{table}[t]
\centering
\caption{
Per-action intra-class distances ($\downarrow$) for visual and tactile
representations. Physical Words yield lower distances than latent-only
supervision across all evaluated actions.
}
\label{tab:within_class_distances}
\small
\setlength{\tabcolsep}{6pt}
\renewcommand{\arraystretch}{1.15}
\begin{tabular}{@{}lcccc@{}}
\toprule
\multirow{2}{*}{\textbf{Action}}
& \multicolumn{2}{c}{\textbf{Visual}}
& \multicolumn{2}{c}{\textbf{Tactile}} \\
\cmidrule(lr){2-3}
\cmidrule(lr){4-5}
& Latent only ($O$) & Physical Words ($WO$)
& Latent only ($T$) & Physical Words ($WT$) \\
\midrule
Pick
& 0.0049 & \textbf{0.0030}
& 0.0003 & \textbf{0.0001} \\
Place
& 0.0050 & \textbf{0.0024}
& 0.0002 & \textbf{0.0001} \\
Approach
& 0.0044 & \textbf{0.0026}
& 0.0003 & \textbf{0.0002} \\
Move
& 0.0062 & \textbf{0.0041}
& 0.0002 & \textbf{0.0001} \\
\bottomrule
\end{tabular}
\end{table}

\begin{figure}[t]
    \centering
    \includegraphics[width=0.7\linewidth]{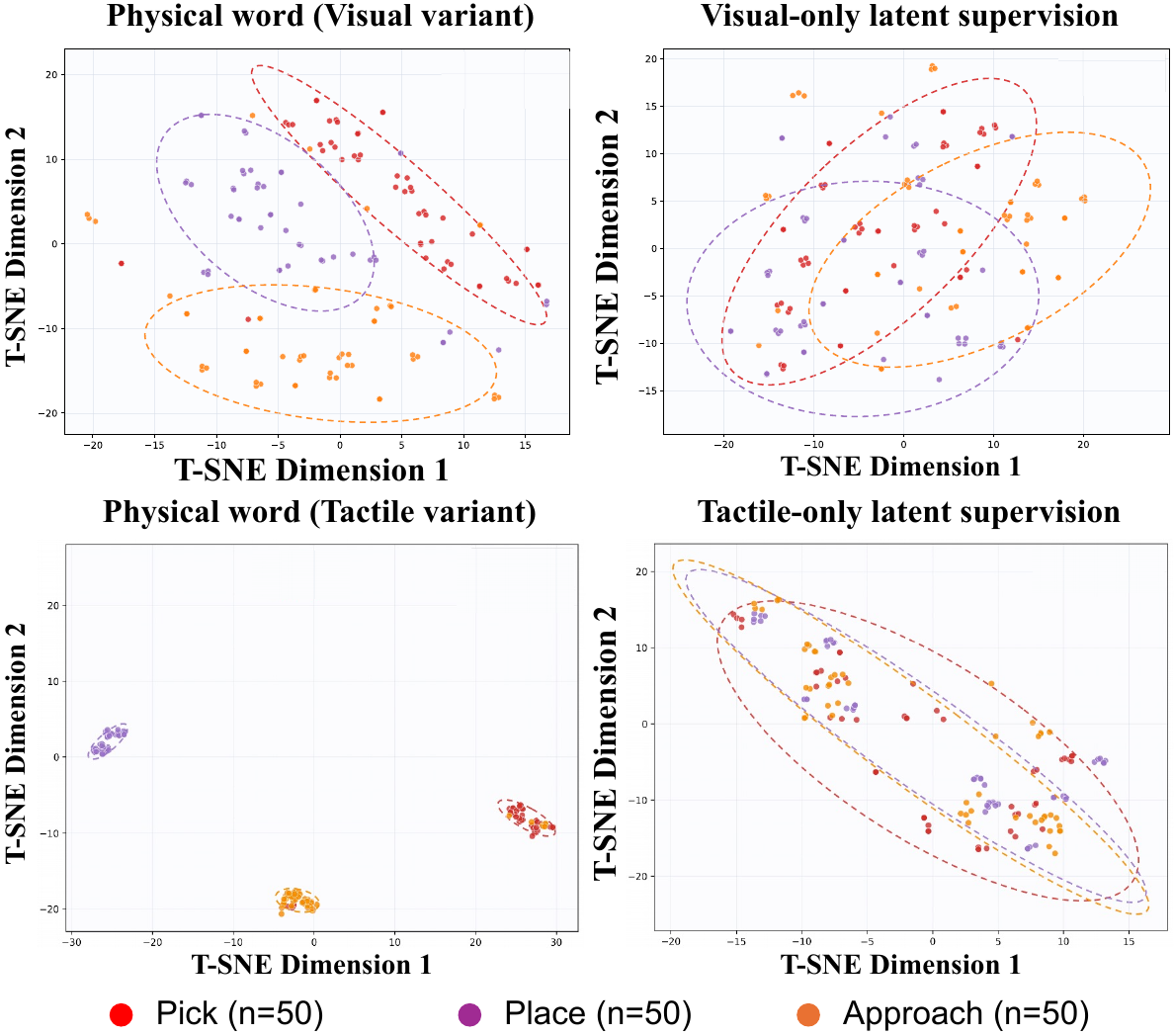}
    \vspace{-0.3cm}
  \caption{\textbf{t-SNE visualization of contextual hidden states.}
We compare Physical Words with latent-only supervision for visual (top) and tactile (bottom) representations. Colors denote atomic actions: \textit{pick}, \textit{place}, and \textit{approach}, with 50 samples per action.}
    \label{fig:tsne}
\end{figure}

\begin{figure}[t]
    \centering
    \includegraphics[width=0.9\linewidth]{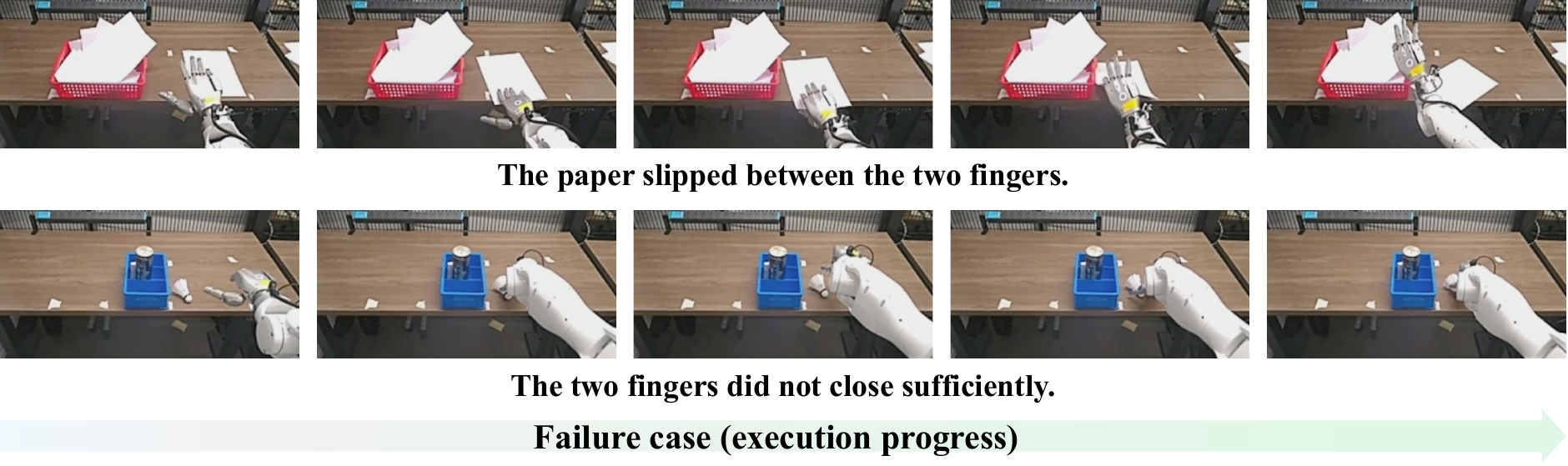}
  \caption{\textbf{Failure case.}
Representative execution failures during contact-rich manipulation.}
    \label{fig:failure}
\end{figure}

\section{Additional Representation Analysis}
\label{app:representation_analysis}
We further examine whether Physical Words learn more semantically structured representations than latent supervision alone. Complementing the attention visualizations and average intra-class distances in the main paper, we report per-action intra-class distances, comparing Physical Word representations with those learned through latent supervision alone. We also use t-SNE to visualize whether representations of the same action cluster across different scenes.

\subsection{Detailed Intra-Class Distance Analysis}
\label{app:within_class_distances}

\textbf{Experimental setup.}
We evaluate \textit{pick}, \textit{place}, \textit{approach}, and \textit{move}, the four most frequent atomic actions in our downstream real-world data. For each action, we randomly sample 50 examples for both visual and tactile analyses. We compare visual Physical Words ($WO$) with visual-only latent supervision ($O$), and tactile Physical Words ($WT$) with tactile-only latent supervision ($T$). Each comparison uses identical samples. We extract the corresponding contextual hidden states and compute the mean pairwise cosine distance within each action class to measure representational consistency across scenarios.

\textbf{Results and analysis.}
Table~\ref{tab:within_class_distances} shows that atomic-word supervision reduces intra-class distances for all four actions in both modalities. Averaged across actions, visual and tactile distances decrease by approximately 41\% and 50\%, respectively. These consistent reductions suggest that shared action labels encourage representations of the same action to remain semantically coherent across varying contexts. Together with future-latent supervision, semantic anchoring helps organize context-dependent physical information around reusable action semantics.

\subsection{t-SNE Visualization of Physical Word Representations}
\label{app:tsne}

We compare Physical Words with latent-only supervision through separate t-SNE analyses of visual ($WO$ versus $O$) and tactile ($WT$ versus $T$) representations. Using samples from Section~\ref{app:within_class_distances}, we project each variant's contextual hidden states into two dimensions and color the points by atomic action.
As shown in Figure~\ref{fig:tsne}, Physical Words form more compact clusters with clearer separation between atomic action categories than latent-only supervision in both modalities. Consistent with the intra-class distance results, these patterns suggest that semantic anchoring promotes shared action semantics across scenes. Moreover, the tactile feedback results in Section~\ref{app:real_tactile_feedback} show that predicted tactile latents can replace sensor-derived inputs with only modest performance degradation. Together, these findings suggest that Physical Words improve semantic organization while retaining the contact information needed for tactile-conditioned control.

\section{Failure Case Analysis}
\label{ap:fca}

Figure~\ref{fig:failure} illustrates two failures in dexterous manipulation: paper slippage and insufficient finger closure.

\textbf{Paper slippage.}
Because the paper is thin, grasping too close to its edge provides limited contact area, allowing it to slip between the fingers during lifting. Future data collection could emphasize more stable grasp locations farther from the paper edge.

\textbf{Insufficient finger closure.}
During shuttlecock grasping, the thumb contacts the object before the index finger achieves sufficient closure. The resulting tactile feedback appears to trigger lifting prematurely, before a stable grasp is established. Despite the initial failure, PhysWord detects that the shuttlecock remains unlifted, predicts the Physical Word \texttt{approach} again, and initiates another grasping attempt. This behavior demonstrates its ability to reuse action primitives for recovery based on execution feedback.

\section{Additional Implementation Details}
\label{app:implementation_details}

\subsection{Baseline Implementation Details}

For UniVTAC, we use baseline results reported in FTP-1~\cite{yuan2026ftp} and N0-VTLA~\cite{team2026n_0} where available and reproduce missing results using official implementations and pretrained checkpoints. All reproduced baselines use the same official training split as PhysWord, with 50 demonstrations per task. We match their image input resolutions to those of PhysWord, as well as tactile input resolutions for tactile-enabled methods, while retaining each method's official fine-tuning configuration. Since FTP-1 uses task-dependent camera configurations on UniVTAC, we use both head and wrist views when reproducing its results on previously unreported tasks. For action execution, we follow FTP-1 by executing only the first action of each predicted chunk before replanning.

For real-world evaluation, we compare against $\pi_{0.5}$ and FTP-1.
$\pi_{0.5}$ is a PaliGemma-based~\cite{beyer2024paligemma} VLA model that uses flow matching for action generation. We initialize from its official pretrained checkpoint and fine-tune the entire model on downstream demonstrations. Following the official implementation, all RGB images are resized to $224\times224$. We retain this resolution to preserve compatibility with the pretrained visual encoder.
FTP-1 extends the VLA model with sensor-specific tactile encoders and a shared tactile expert that conditions the action expert. We initialize from its released pretrained checkpoint and follow its downstream fine-tuning protocol. Tactile inputs are resized to $224\times224$ and processed by a sensor-specific Transformer followed by a shared T3 Transformer module. We retain its normalization scheme for proprioceptive states and actions, as well as its optimizer and associated hyperparameters.
For a controlled comparison, all methods use the same task demonstrations, camera views, and total number of training iterations. We evaluate the final checkpoint of each method.

\begin{figure}[t]
    \centering
    \includegraphics[width=0.8\linewidth]{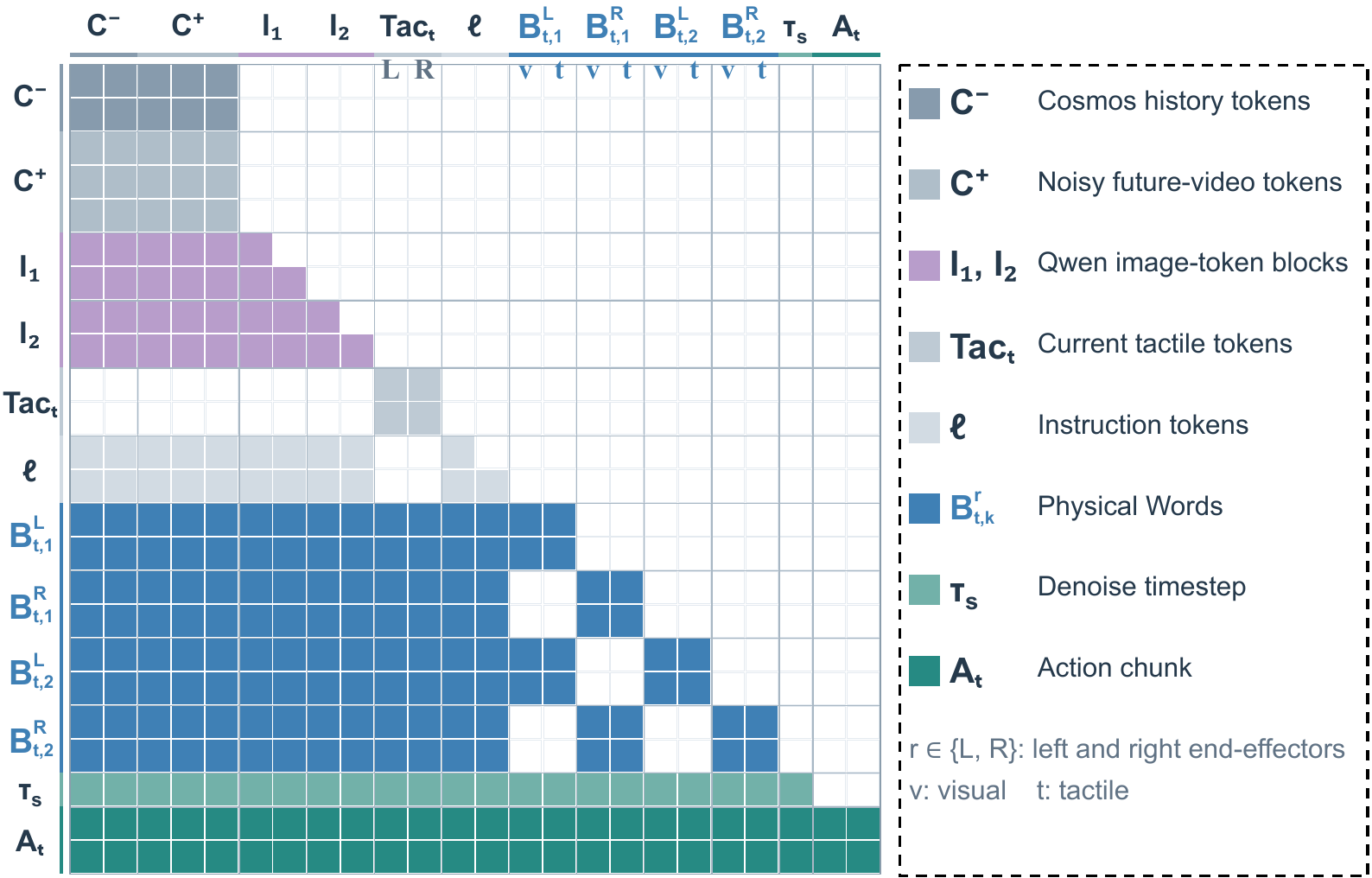}
    \caption{Attention mask for the reasoning-and-action branch. Rows denote queries and columns denote keys. Colored cells indicate allowed attention, while white cells indicate masked connections.}
    \label{fig:apattention_mask}
\end{figure}

\subsection{PhysWord Implementation Details}

\textbf{Training configuration.}
For supervised fine-tuning, we train PhysWord for 100 epochs using AdamW with zero weight decay and a global batch size of 48 on NVIDIA H100 GPUs. The base learning rate is $10^{-4}$, with reduced rates of $2\times10^{-6}$ for the Cosmos backbone and $5\times10^{-5}$ for the T3 tactile encoder. We apply linear warmup over the first 5\% of optimizer steps, followed by cosine decay to zero, using BF16 mixed precision. 
To stabilize tactile latent alignment, we apply stop-gradient to the future tactile targets, blocking gradients from the tactile cosine distance loss to T3 through these targets. T3 remains trainable through the current tactile input pathway, while FTP-1 initialization and the reduced learning rate help preserve its pretrained representations and limit target drift.
Each action dimension is normalized using the task-specific training-set 1st and 99th percentiles and clipped to $[-1,1]$. Following FTP-1, tactile RGB inputs are resized to $224\times224$.

\textbf{Input processing and attention.}
Head-view images supplied to the temporal perception branch are resized to $256\times256$, while the remaining views use Qwen3-VL's native adaptive-resolution processing. Attention patterns for image and text inputs follow the respective pretrained configurations of Cosmos and Qwen3-VL. Figure~\ref{fig:apattention_mask} supplements the main-text description with the complete attention mask, including the preceding observation context and action tokens. Token counts in the schematic are reduced for readability.

\textbf{Inference and real-time chunking.}
For real-world deployment, we adopt real-time chunking (RTC)~\cite{black2026real} to overlap policy inference with action execution. Each predicted chunk contains $H=16$ actions, eight of which are executed before switching to the next chunk. During execution, the policy asynchronously predicts the next chunk using newly acquired observations, allowing the robot to continue executing the current action sequence without pausing for inference.

\subsection{Inference Efficiency and Runtime Optimization}
\label{app:inference_efficiency}

All measurements follow the inference configuration in Section~\ref{sec:simulation} and run on a single NVIDIA H100 GPU with batch size one. Starting from a baseline implementation running at 4.9 Hz, we apply the following runtime optimizations. None of them modifies model weights, attention masks, or the inference configuration, and all observation-dependent features are recomputed for every request.
Table~\ref{tab:inference_latency} reports the cumulative latency
improvements in the order in which the optimizations are applied.

\textbf{Full CUDA Graph capture.}
The baseline captures only the Physical Word and action-denoising steps as CUDA graphs. We extend graph capture to all major forward passes, including the RGB and tactile encoders, the history VAE, the Cosmos denoising step, and the context prefill. For each request, new inputs are copied into static buffers before graph replay, reducing Python dispatch and kernel-launch overhead.
As shown in Row~1 of Table~\ref{tab:inference_latency}, full CUDA Graph capture reduces median latency from 205.4 to 163.2\,ms, a reduction of 42.2\,ms.

\textbf{Mask-preserving FlashAttention.}
Several attention operations in the reasoning-and-action branch originally relied on explicit dense masks, which precluded efficient attention kernels. We verify the structure of each mask and dispatch it to an equivalent FlashAttention path. Causal regions over the context and Physical Words use bottom-right-aligned causal attention. Tactile queries attend bidirectionally to all keys within the tactile-token block. The timestep and action queries of each denoising step are computed separately according to their respective visibility. Masks that do not match a verified pattern fall back to the original implementation, preserving the causal and tactile-isolation constraints.
Row~2 shows that mask-preserving FlashAttention further reduces
median latency from 163.2 to 145.5\,ms, saving 17.7\,ms.

\textbf{GPU RGB/tactile preprocessing.}
Raw RGB images are transferred asynchronously through pinned memory. A Triton kernel then performs normalization and Qwen3-VL patch packing directly on the GPU, producing inputs that match those of the original processor. 
This reduces median latency by 13.8\,ms
(Row~3).
Furthermore, tactile images from all sensors and history frames are resized and normalized in a single batched GPU operation, which is also captured as a CUDA graph.
This provides an additional
3.9\,ms reduction when introduced later in the optimization
sequence (Row~8). 
Together, these two steps account for
17.7\,ms of the total latency reduction.

\textbf{Fused RMSNorm, packed projections, and fused RoPE.}
We implement custom Triton kernels for RMSNorm and rotary position embedding (RoPE), fusing their elementwise and reduction operations to reduce intermediate tensors and memory traffic.
The fused RMSNorm replaces the RMSNorm layers in both the Qwen and Cosmos backbones, retaining FP32 reduction and the original casting order while reducing median latency by 15.2\,ms (Row~4) and 2.9\,ms (Row~6), respectively. 
In the reasoning-and-action branch, the query, key, and value projections, as well as the gate and up projections of each MLP, are packed into single GEMMs; together with fused RoPE, this reduces median latency by another 8.2\,ms (Row~5). These three steps collectively contribute a latency reduction of 26.3\,ms.

\textbf{Session/Positional Metadata caching.}
For a fixed instruction and input shape, instruction tokens, attention masks, visual grid information, and positional and rotary metadata remain unchanged across requests. We cache these after the first call and reuse per-layer KV workspaces within each request, while RGB, tactile, and head-view history inputs are updated at every step.
Specifically, session metadata caching and positional metadata caching
reduce median latency by 3.7\,ms (Row~7) and 1.5\,ms
(Row~9), respectively, contributing a combined reduction
of 5.2\,ms.

\textbf{VAE/RGB stream overlap.}
Encoding the head-view history with the video VAE is independent of encoding the current frames. We therefore run the VAE on an auxiliary CUDA stream concurrently with the RGB and tactile encoders. A stream dependency ensures that the Cosmos step starts only after the history latents are ready.
This reduces median latency
from 96.3 to 94.6\,ms (Row~10), saving another 1.7\,ms.
With all optimizations enabled, inference runs at 10.6\,Hz,
corresponding to a $2.2\times$ speedup over the baseline.

\begin{table}[t]
\centering
\caption{End-to-end inference latency on a single NVIDIA H100 GPU as runtime optimizations are applied cumulatively. Frequency is computed from median latency.}
\label{tab:inference_latency}
\small
\setlength{\tabcolsep}{6pt}
\begin{tabular}{clccc}
\toprule
Row & Implementation & Median (ms) & P95 (ms) & Freq. (Hz) \\
\midrule
 & Baseline                            & 205.4 & 216.5 & 4.9 \\
1 &+ Full CUDA Graph capture           & 163.2 & 164.0 & 6.1 \\
2 &+ Mask-preserving FlashAttention    & 145.5 & 146.3 & 6.9 \\
3 &+ GPU RGB preprocessing             & 131.7 & 132.0 & 7.6 \\
4 &+ Fused RMSNorm (Qwen)              & 116.5 & 119.0 & 8.6 \\
5 &+ Packed projections and fused RoPE & 108.3 & 108.7 & 9.2 \\
6 &+ Fused RMSNorm            & 105.4 & 105.8 & 9.5 \\
7 &+ Session metadata caching          & 101.7 & 102.7 & 9.8 \\
8 &+ GPU tactile preprocessing         &  97.8 &  99.4 & 10.2 \\
9 &+ Positional metadata caching       &  96.3 &  97.9 & 10.4 \\
10 &+ VAE/RGB stream overlap            & \textbf{94.6} & \textbf{95.1} & \textbf{10.6} \\
\bottomrule
\end{tabular}
\end{table}

\section{Real-World Evaluation Protocol}
\label{app:evaluation_protocol}

\subsection{Task Definitions and Success Criteria}
We mark 30 distinct positions on the table and evaluate each task with the manipulated object's center placed at each position in turn. Target receptacles or fixtures, including the basket, shuttlecock tube, express box, and toy-car base, are also randomly initialized within a task-specific region. All methods are evaluated under matched initial configurations, with the environment reset after each trial. Success is assessed manually using the task-specific criteria below. Representative task executions are shown in Figure~\ref{fig:vis} of the main paper and the supplementary videos.

\textbf{Pick A4 Paper.}
An A4 sheet is initially placed entirely on the tabletop, with no edge overhanging. The dexterous hand first establishes contact with its five fingers and uses friction to gradually slide the sheet toward the table edge. Once part of the sheet extends beyond the edge, the hand grasps and lifts it, then transfers it into a basket. A trial succeeds only if the entire sheet is placed inside the basket.

\textbf{Shuttlecock Insertion.}
One to three shuttlecocks are randomly placed on the table. The robot must grasp each shuttlecock by its feathers without damaging it, insert it into a shuttlecock tube, and press it down with the palm. Success requires all shuttlecocks to be fully inserted and remain intact. Partial insertion does not count as success.

\textbf{Express Packing.}
The robot places an object from the tabletop into an express box and closes all four flaps. It then picks up tape with the right gripper and coordinates both grippers to apply the tape along the lid seam and tear it off. Success requires completion of the entire sequence and a fully sealed box. Tape that fails to secure the lid seam is counted as a failure.

\textbf{Powered Screw Driving.}
The left gripper grasps a screw and positions it in the designated hole on a toy-car base, while the right gripper operates a powered screwdriver to drive it into place. A trial succeeds only when the screw is fully driven into the designated hole.

\begin{figure}[t]
    \centering
    \includegraphics[width=0.9\linewidth]{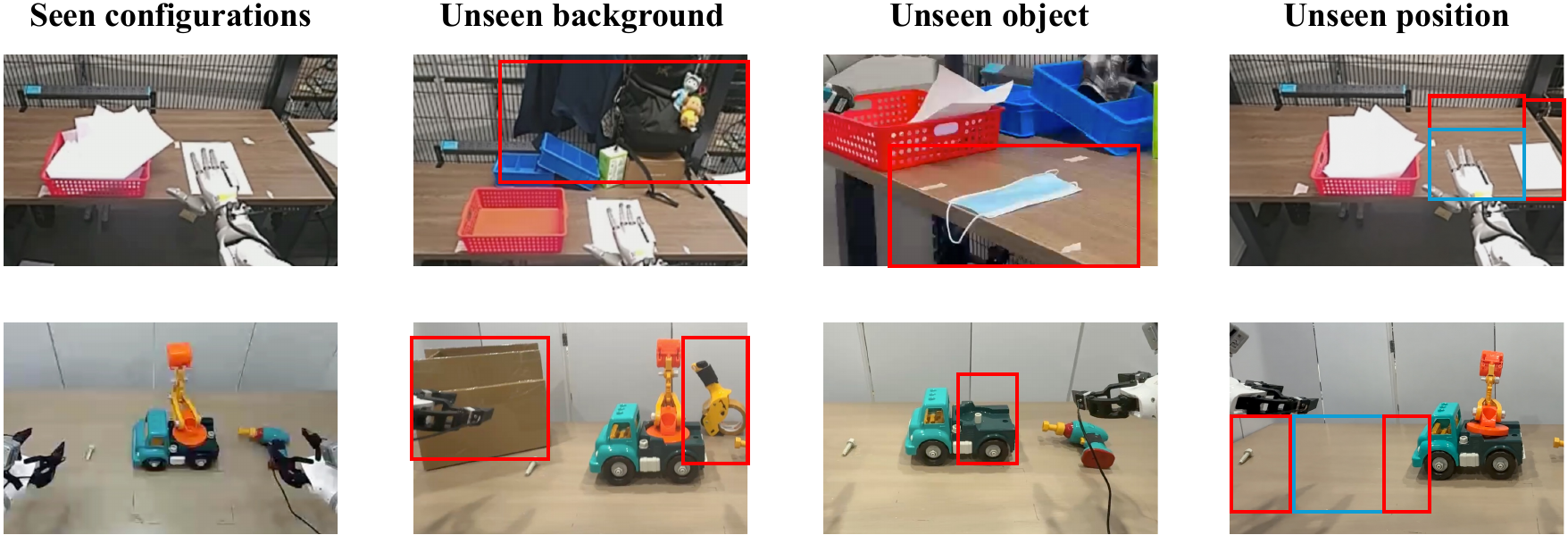}
  \caption{\textbf{Real-world generalization.}
Examples of seen configurations and unseen variations in backgrounds, objects, and positions. Blue borders indicate configurations within the training distribution, while red borders indicate unseen configurations.}
    \label{fig:ap_unseen}
\end{figure}

\subsection{Evaluation under Unseen Configurations}
We evaluate generalization on \textit{Pick A4 Paper} and \textit{Powered Screw Driving} under three types of distribution shift: unseen backgrounds, object instances, and object positions. Each factor is varied separately while the remaining conditions are controlled. All methods are tested on the same configurations using the success criteria defined above. Figure~\ref{fig:ap_unseen} illustrates these settings.

\textbf{Unseen backgrounds.}
We introduce background distractors absent from the training demonstrations in both tasks, while retaining the task objects and their placement distribution.

\textbf{Unseen objects.}
For \textit{Pick A4 Paper}, we replace the A4 sheet with napkins, masks, notebooks, tennis balls, and sheets of different colors and sizes. The objective remains to pick up the object and place it entirely inside the basket, allowing the grasping strategy to vary with the object. For \textit{Powered Screw Driving}, we use toy-car bases with unseen structural configurations, such as variants without wheels or the front section, while retaining the screw-fastening objective.

\textbf{Unseen positions.}
We place the manipulated objects at workspace positions not represented in the training demonstrations, while retaining the training backgrounds and object instances.

\end{document}